\documentclass[letterpaper, 10 pt, conference]{ieeeconf}  

\IEEEoverridecommandlockouts                              

\usepackage{cite}
\usepackage{amsmath,amssymb}
\usepackage{graphicx}
\usepackage{booktabs}
\usepackage{xcolor}
\usepackage[utf8]{inputenc}
\usepackage{newunicodechar}
\usepackage[linesnumbered,ruled,vlined,noend]{algorithm2e}
\makeatletter
\let\NAT@parse\undefined
\makeatother
\usepackage[hidelinks]{hyperref}

\newunicodechar{，}{,}

\SetKwInOut{Input}{Input}
\SetKwInOut{Output}{Output}
\SetKwInOut{State}{State}
\SetInd{0.2em}{0.5em}
\SetKwIF{If}{ElseIf}{Else}{if}{then}{else if}{else}{end}
\SetKwFor{For}{for}{do}{end}
\SetKwFor{While}{while}{do}{end}
\SetAlFnt{\normalsize}
\SetAlCapFnt{\small}
\SetAlgoNlRelativeSize{0}

\title{\LARGE \bf Receding-Horizon Pushing with Composable Object-Centric Policies}

\author{ 
Zhiyi~Yuan， 
Tianrun~Hu, 
Anxing~Xiao,  
Yuhong~Deng, 
David~Hsu, 
Hanbo~Zhang 
\thanks{Zhiyi Yuan, Tianrun Hu, Anxing Xiao, Yuhong Deng, and David Hsu are with the School of Computing and Smart Systems Institute, National University of Singapore, Singapore. 
Hanbo Zhang is with the Shanghai Innovation Institute, Shanghai, China and the School of Computing and Smart Systems Institute, National University of Singapore, Singapore.
Correspondence to: yuanzy9920@gmail.com, zhanghanbo@sii.edu.cn.}
} 

\begin{document}
\bstctlcite{IEEEcontrol}

\maketitle
\thispagestyle{empty}
\pagestyle{empty}

\begin{abstract}

Non-prehensile manipulation is practical for relocating large, heavy, or geometrically ungraspable objects.
Yet, long-horizon pushing of arbitrarily-shaped 3D objects couples three problems: 1) where to push the object so as to approach the target pose, 2) whether each push is stable and
reachable, 3) whether subsequent actions remain feasible.
We present an object-centric pushing policy within a feedback-guided hierarchical framework.
At the low level, a learning-based policy predicts contact actions from a pose- and scale-normalized point cloud, conditioned on a near single-step subgoal.
A stability score is applied to evaluate the predicted contacts by a quasi-static sliding-versus-tipping analysis.
At the high level, BIT$^*$ first searches for an object path, and the next several subgoals are checked by contact prediction and robot motion planning for future feasibility.
Failed motion plans, as feedback, change the local path costs and trigger re-planning.
During execution, only the first feasible action is executed. In simulation, we evaluate 22 objects in six different scenes, upon which we also conduct comprehensive ablation studies.
Results demonstrate that our method outperforms baselines with a clear margin and can reliably achieve long-horizon object pushing tasks under different situations.
We also report quantitative real-robot experiments with a Franka arm and qualitative demonstrations with a mobile manipulator for large and heavy objects, with directly zero-shot sim-to-real transfer.
The project page is available at: https://yzy14606.github.io/Receding-Horizon-Pushing-Website/.

\end{abstract}

\section{INTRODUCTION}

Pushing is practical when an object is too large or too heavy for grasping.
Planar pushing of simple objects is well studied \cite{mason1986scope,lynch1992manipulation,akella1998posing}, whereas practical objects
may have irregular 3D geometry and can tip or roll.
For such objects, a contact that produces a desired planar push also depends on the object's support geometry and force direction.
Moreover, the error compounds over a long horizon.
For example, moving along a collision-free object path may require robot contacts that are unreachable, and an inaccurate push may leave the object in a corner or near clutter where future actions are infeasible. These challenges expose a gap between local pushing policies and long-horizon pushing tasks.
Analytical and model-predictive methods provide explicit physical reasoning but usually rely on precise and expensive contact models \cite{10.3389/frobt.2020.00008,mericli2015push,bui2026pushanythingsinglemultiobject}.
Learned object-centric policies get rid of contact modeling and reduce computation, but a local policy does not account for robot kinematics, scene collisions, or the feasibility of future actions
\cite{zhou2024hacmanlearninghybridactorcritic,doi:10.1177/02783649241273668,9811645,
zhu2023learning,10801782,wu2022vat,chenobject}. 
Therefore, a question arises: \textit{how can we
achieve generalizable learned contacts and compose them for long-horizon pushing reliably?}

\begin{figure}[t]
	\centering
	\includegraphics[width=\linewidth]{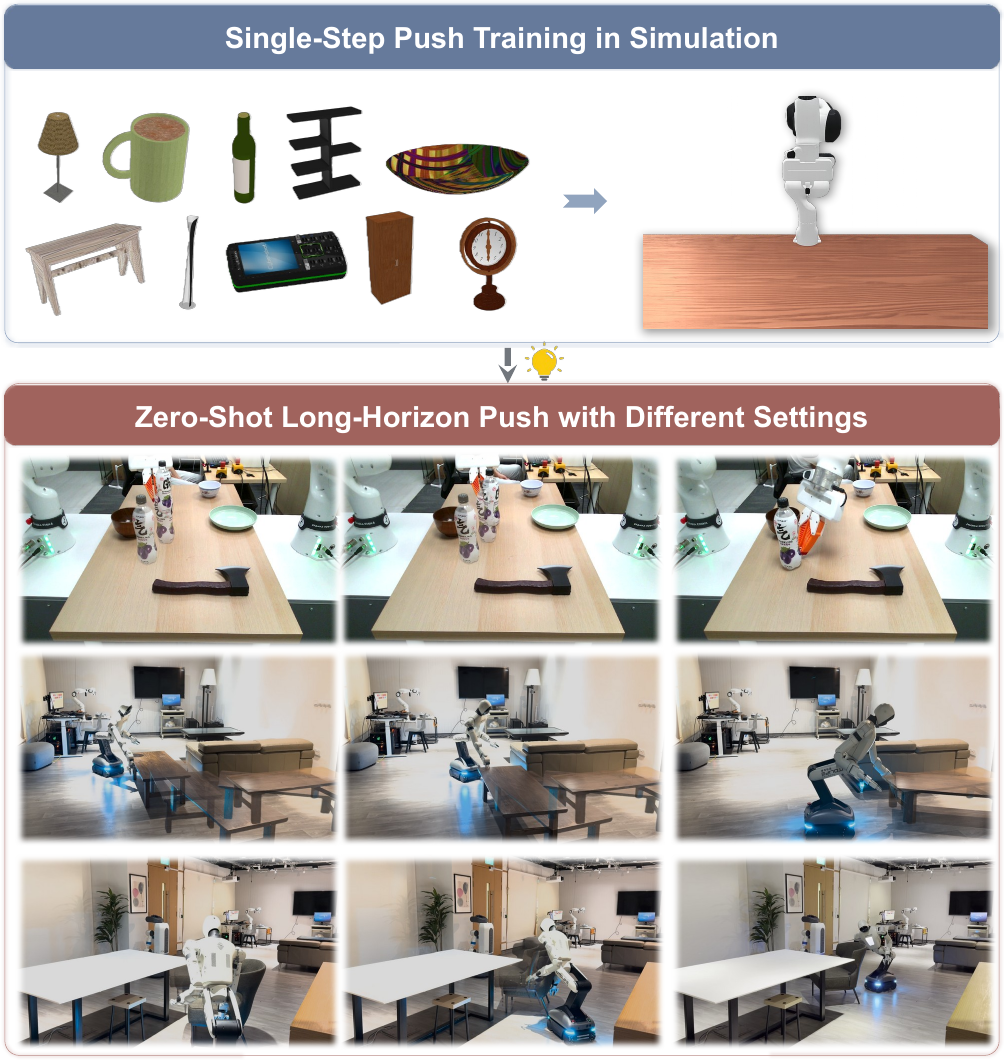}
	\caption{A single-step contact policy is composed with stability scoring and
    feedback-guided hierarchical planning for long-horizon deployment.}
	\label{fig:first-figure}
    \vspace{-0.7cm}
\end{figure}

To do so, our key idea is that a push is a contact on the object, so the action should be directly associated with the geometry of the object itself, which can be learned inside the object-canonical frame. Simultaneously, object stability and long-horizon feasibility, subject to environmental constraints, should be addressed by explicit, model-based verification rather than left to the learned policy.
We therefore formulate the low-level pushing actions in the object frame normalized for pose and scale.
In this frame, the map from a desired pose change to a contact is learned for each object shape, so a network trained on a small dataset, collected from a single push direction, generalizes to unseen poses, scales, and objects.
Stability is checked explicitly by a quasi-static analysis of the object's support polygon, which softly rejects contacts that would tip the object.
This stability test is scale-invariant and consistent with the normalized policy.
Finally, a bi-level planner, considering both object trajectories and robot motions, shapes the object's path to keep future pushes reachable and feasible, with closed-loop replanning addressing model inaccuracy.

We evaluate 22 ShapeNet \cite{shapenet2015} objects in six simulated representative scenes and report quantitative
real-world tabletop experiments on a Franka arm.
The result concludes that a policy trained on only ten objects, each pushed from a single direction, moves unseen, geometrically complex objects to their goals.
Long-horizon pushing results demonstrate that our policy is composable for bi-level planners, and can effectively execute long-horizon tasks reliably.
We additionally show a qualitative deployment on
a mobile manipulator pushing large furniture, showing zero-shot generalization to different robots and task settings.

Our contributions include: 
(1) An object-centric learning approach for generalizable and composable pushing of geometrically complex 3D objects, which enables zero-shot transfer across embodiments, unseen objects and environments;
(2) A feedback-guided bi-level planner integrated with the object-centric policy, which enables reliable long-horizon object pushing.

\section{Related Work}

\subsection{ Object-Centric Manipulation Policies }
Object-centric representations have been used to improve
the transfer of robotic manipulation policies across scenes and tasks \cite{Yu_2025_ICCV,zhu2023learning}.
Existing methods enhance the policy's perception of task-relevant objects
\cite{10609992,yuan2022sornet,Li_2024_CVPR,10610131,10.3389/fnbot.2025.1585386}
or isolate the object to learn its representation independently~\cite{wu2022vat,chenobject,Pan_2025_CVPR}.
However, these policies may still be coupled with scene- or robot-specific configurations, 
or suffer from large spatial variations caused by different object pose. 
To address these limitations, our method learns goal-conditioned contacts in an object-centric canonical space that normalizes both object pose and scale variations, decoupling the policy from robot- and scene-specific configurations and yielding composability and generalization with high data efficiency.

\subsection{Object Pushing }
Early methods established the foundation for model-based pushing through quasi-static analysis,
contact modeling, and related techniques \cite{mason1986scope,lynch1992manipulation,akella1998posing}.
Recent model-based methods use online optimization and feedback replanning
\cite{bui2026pushanythingsinglemultiobject,10556602,11176447,9811942}.
However, these methods mostly focus on 2D objects and do not explicitly account for full 3D geometry.
Learning-based methods have been explored for 3D object pushing.
Some recent works~\cite{zhou2024hacmanlearninghybridactorcritic} model objects as 3D
point clouds for non-prehensile manipulation, but focus on contact-rich 6D pose adjustment
rather than coupling contact prediction with future robot-motion feasibility.
Other methods consider the toppling stability of 3D object pushing, but use geometric approximations most directly applicable to boxes, cylinders, or extruded
shapes \cite{11128166,11246770}.
We focus on the practical long-horizon pushing tasks, considering both geometrically complex 3D object stability and long-horizon feasibility by integrating a receding-horizon planner with a composable low-level pushing policy.

\subsection{Hierarchical Manipulation Planning}
Hierarchical manipulation is an important strategy to address long-horizon tasks
\cite{triantafyllidis2023hybrid,10611125,xie2023hierarchical,xiao2026hume}.
Typical hierarchical methods decompose long-horizon tasks into high-level subgoals
and execute them with low-level planners or controllers
\cite{Li2018PushNet,7139389,nasiriany2022augmenting}.
However, their high-level modules operate without considering low-level execution previously.
Bi-level planning considers the trajectory feasibility of the complete task before action execution \cite{bilevel-planning-blog-post,Li2025RSS}.
However, standard bi-level planners typically assume known dynamics and use feasibility mainly to reject or rank high-level candidates, rather than exploiting failures to adapt the high-level planner, which is difficult to model psuhing in practice.
In contrast, we integrate a high-level planner that generates subgoals and preserve space for future pushes with a learned goal-conditioned low-level policy that executes them. Low-level feasibility failures are fed back to update the high-level cost distribution, enabling robust receding-horizon push planning without requiring accurate dynamics.

\section{Problem Formulation}
\label{sec:formulation}


We consider the problem of a rigid 3D object that is moved by planar pushes on a horizontal support surface with static obstacles.
The inputs are its initial point cloud $P_0$, $X_0\in SE(2)$, target pose $X_g\in SE(2)$, and scene geometry $\mathcal{W}$, which is reconstructed from perceptions.
The output is an object-pose sequence $X_{0:K}$ and a push sequence $u_{0:K-1}$, where $K$ is determined by the planner and $u_k=(p_k,v_k,d_k)$ contains a surface contact point, a push direction, and a push distance.
For every $k$, the contact must satisfy $u_k=\pi(X_k,X_{k+1},P_k)$, $u_k\in\mathcal{S}(X_k,P_k) \cap \mathcal{R}(X_k,\mathcal{W})$.
Here, $P_k$ is the estimated object point cloud at $k$, $\pi$ is the pushing policy that associates each desired single-step pose transform with a contact on $P_k$ achieving it, $\mathcal{S}(X_k,P_k)$ collects the contacts that do not tip the object at pose $X_k$, and $\mathcal{R}(X_k,\mathcal{W})$ collects the contacts the robot's end-effector can reach at $X_k$ without colliding with $\mathcal{W}$.
We assume known robot kinematics, a static scene during pushing, and single-point end-effector contact.
The stability model further uses a quasi-static
approximation and estimates the support polygon and center of mass from the observed geometry.

\begin{figure*}[ht]
	\centering
    \includegraphics[width=0.95\textwidth]{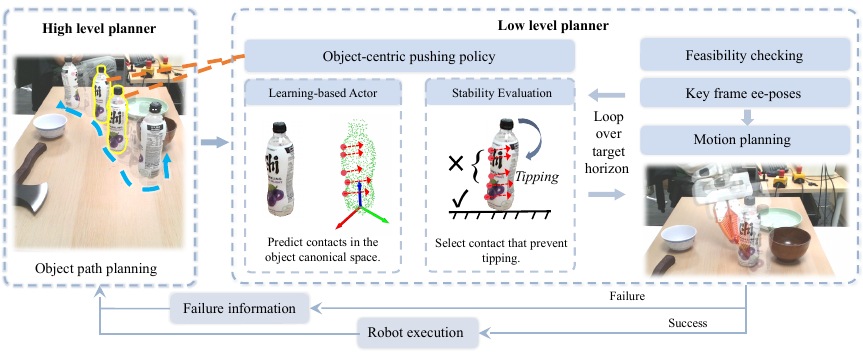}
    \vspace{-0.3cm}
	\caption{Method overview. The high level plans an object path, the low level
    checks contact and robot-motion feasibility, and a failed check updates the high level cost distribution before replanning.}
	\label{fig:overview}
    \vspace{-0.6cm}
\end{figure*}

\section{Method}

Our method combines three components: an object-centric contact-based policy, a quasi-static stability score $\tau$, and a feedback-guided hierarchical planner.
The high-level state is the planar object pose $X_k\in SE(2)$. The policy consumes a canonical point
cloud $P_k^{(\phi)}$ and a canonical desired transform $T_k$ (Sec.~\ref{subsec:policy}).
At execution step $k$, the high-level planner generates an object path from $X_k$ to the goal.
For at most the next $h$ path segments, the low level predicts contacts, evaluate them with $\tau$, selects the best for each pushing step, and plans the associated robot motions. 
If all checked steps are feasible, the first pushing action is executed.
The object is then perceived again, and the loop repeats until the target pose is reached. Figure~\ref{fig:overview} demonstrates the pipeline.

\subsection{High-Level Object Path Planning}
\label{subsec:planner}

\subsubsection{Object Path Planner}
We leverage BIT$^*$~\cite{doi:10.1177/0278364919890396} as the high-level object planner.
At step $k$,  BIT$^*$ searches for a collision-free
pose sequence $X_{k:K}$ for the target object.
Let $D(X_i,X_{i+1})\geq0$ denote the $SE(2)$ edge
metric and $C(X_i,X_{i+1})\geq0$ an adaptive edge penalty
initialized to zero at the start of a task. BIT$^*$ minimizes
\begin{equation}
    J(X_{k:K})=\sum_{i=k}^{K-1}
    \bigl[1+C(X_i,X_{i+1})\bigr]D(X_i,X_{i+1}).
    \label{eq:path-cost}
\end{equation}
Thus, before any low-level planning or motion failures, the objective is the usual path cost.
Feedback changes the cost of path segments softly by updating the cost term $C(X_i,X_{i+1})$ after each failure.
Besides, we inject clearance preference into the search prior. 
The planner first tries to search the object path with a minimum $5$ cm distance to any obstacles.
If it fails or does not return a path within time limits, the distance decreases until a feasible path is found.

\subsubsection{Future Feasibility Checking and Feedback}
At time step $k$, the contact policy proposes candidates for the desired moves in the next $h$ steps.
The stability score contributes to selecting the best one for each step, and the robot motion planner checks motion feasibility from its corresponding pre-push and post-push keyframes.
If a motion cannot be found within a fixed planning
time, the candidate object path is rejected. Denote $\bar X$ as the $xy$-position on the path at which the planner feasibility check fails. 
The corresponding 2D transition is denoted by $\bar e$.
Their nearby edges are penalized.
The penalty has two components: one decays with planar distance from the failed waypoint, while the other decays with angular difference and is further modulated by distance. Both components use truncated Gaussian decay, and accumulate
across replans:
\begin{equation}
    C(e) \leftarrow C(e)
    +5K_{\mathrm p}(e;\bar X)
    +5K_{\mathrm t}(e;\bar e),
    \label{eq:cost-update}
\end{equation}
The kernels are
$K_{\mathrm p}=e^{-d_{\mathrm p}^{2}/(2\sigma_{\mathrm p}^{2})}$
and
$K_{\mathrm t}
=e^{-(d_{\mathrm t}^{2}/\sigma_{\mathrm p}^{2}
+d_{\theta}^{2}/\sigma_{\theta}^{2})/2}$
for $d_{\mathrm p}\lesssim3\sigma_{\mathrm p}$, and zero otherwise.
Here, $d_{\mathrm p},d_{\mathrm t}$ are the planar distances from
$X$ to $\bar X$ and the segment $\bar e$.
The angle $d_{\theta}$ is between the planar directions of
$e$ and $\bar e$; $\sigma_{\mathrm p},\sigma_{\theta}>0$
set the spatial and angular scales.
The high level then replans with Eq.~\eqref{eq:path-cost}.

If all $h$ checks succeed, the robot executes only the first push, observes the new object pose, and replans. The procedure terminates when the goal is reached
or the fixed replanning limit is exhausted.
Notably, an important planner-specific parameter is
$h$, which affects how far the robot is seeing before each physical action.
We ablate the selection of $h$ in Sec. \ref{sec:h_selection}.

\subsection{Low-Level Object-Centric Pushing Policy}
\label{subsec:policy}

\subsubsection{Object Frame Canonicalization}

At the beginning of each task, an object-canonical frame is constructed from the
world-frame point cloud $P_0$.
We assume a known gravity direction.
The $z$-axis of the canonical space passes through the centroid of $P_0$ and is oriented opposite to gravity.
To determine the horizontal axes,
$P_0$ is projected onto the support plane, and PCA is performed on the projected points. The dominant principal direction is defined as the $x$-axis, while the $y$-axis is determined according to the right-hand rule.
Once this frame is established, the normalization length $\ell$ is the maximum
Euclidean distance from its origin $o_0$ to the observed object surface.

 Because a PCA axis has two possible signs, the sign chosen at task initialization is
held fixed by pose tracking. At each step, let $P_k$ be the observed world-frame point cloud. Pose tracking gives the world-frame pose $Y_k=(Q_k,o_k)\in SE(3)$ of the canonical frame, where $Q_k\in SO(3)$. The normalization length $\ell$ is obtained from the origin $o_k$ and $P_k$. Combining the desired next
planar pose $X_{k+1}$ with the tracked roll, pitch, and support height defines
$Y_{k+1}\in SE(3)$. Let $R_k\in SO(3)$ and $r_k\in\mathbb{R}^3$ be the rotation
and translation of $Y_k^{-1}Y_{k+1}$, and let the superscript $(\phi)$ denote the
canonical mapping. The two policy inputs are 
\begin{equation}
    P_k^{(\phi)}=Q_k^T(P_k-o_k)/\ell,\qquad
    T_k=\begin{bmatrix}R_k&r_k/\ell\\\mathbf{0}^{\top}&1\end{bmatrix}.
    \label{eq:canonical-inputs}
\end{equation}
For a predicted canonical contact $u=(p,v,d)$, the corresponding world-frame contact
is $(Q_kp\ell+o_k,\,Q_kv,\,d\ell)$. This construction analytically removes the tracked rigid pose and scalar
length from the policy input, making it generalizable regardless of object poses and scales.

\subsubsection{Pushing-Action Representation}
\label{subsubsec:action-rep}

A single-step push is represented in the canonical frame as contact $u=(p,v,d)$,
where $p\in P_k^{(\phi)}$ is a contact point on the object surface, $v\in\mathbb{S}^2$ is a unit push direction, and $d\in\mathbb{R}_+$ is the push distance.
Our policy predicts it densely over the observed geometry, generating at every point $p_i\in P_k^{(\phi)}$ a contact probability $\hat a_i\in[0,1]$, a direction vector $\hat v_i\in\mathbb{R}^3$, and a distance $\hat d_i\in\mathbb{R}_+$. The direction vector is normalized during training.
The predicted contacts remain explicit, so the stability score and robot-motion planner can evaluate them before execution. The canonical representation reduces pose and scale variations while decoupling camera configurations and robot embodiments from the policy.

\subsubsection{Data Generation}
\label{subsubsec:data-gen}

Each training example is a tuple $(P^{(\phi)},\,T,\,\{(a_i,v_i,d_i)\}_{i=1}^N)$ in the canonical frame.
The inputs are the canonical object cloud $P^{(\phi)}=\{p_i\}_{i=1}^N\subset\mathbb{R}^3$ with $N=1024$
and the canonical single-step transform $T$,
and the per-point targets are a binary contact label $y_i\in\{0,1\}$,
a push direction $v_i$, and a push distance $d_i$,
with $v_i$ and $d_i$ defined on the positive points $\{i:a_i=1\}$.

Examples are collected in ManiSkill3~\cite{taomaniskill3} with random pushing actions.
Per rollout, an object is placed at a fixed position with random yaw, three RGB-D cameras reconstruct its cloud,
and the end-effector pushes it along a fixed direction with a random distance, then retracts while it settles.
From the world-frame object poses before and after the push and the
recorded contact trajectory, we extract the single-step transform $T$ and distance $d$.
The five points nearest each recorded contact are labeled positive ($a_i=1$, inheriting $v$ and $d$) and the rest negative.
The cloud is reduced to $N=1024$ points by farthest-point sampling with the true contacts retained, then mapped to the canonical frame by $\phi$.
Rollouts with no object motion or incomplete reconstruction are discarded.

The main training set consists of 100 objects with 500 examples each, totaling 50,000 examples. A 10-object set with 5,000 examples per object, also totaling 50,000 examples, is used for ablation studies.
Both are fivefold-augmented by canonical-frame jitter
(random $x,y$ translation within the object's convex hull and random yaw)
to tolerate frame-estimation error at test time.

\begin{figure}[t]
	\centering
	\includegraphics[width=\linewidth]{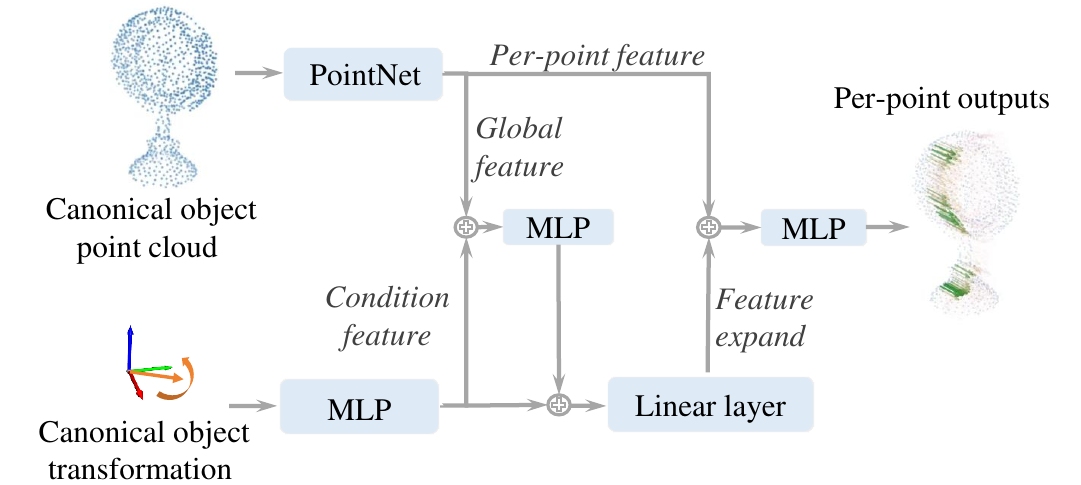}
    \vspace{-0.6cm}
	\caption{Pushing policy network.}
	\label{fig:nn-architecture}
    \vspace{-0.8cm}
\end{figure}

\subsubsection{Pushing Policy}

The policy operates entirely in the canonical frame.
It is a point-wise network that maps the input pair $(P_k^{(\phi)},\,T_k)$
to the per-point outputs $(\hat a_i,\hat v_i,\hat d_i)$.
A PointNet~\cite{qi2017pointnet} backbone of three $1\!\times\!1$ convolution layers
with group normalization and ReLU extracts per-point and global features from
$P_k^{(\phi)}$. A separate MLP embeds $T_k$. The global and transform features are
fused, expanded to all points, and concatenated with the per-point features, as shown
in Fig.~\ref{fig:nn-architecture}. A shared decoder MLP maps each fused feature through
three heads, producing a contact logit passed through a sigmoid to give $\hat a_i$, a direction vector $\hat v_i$, and a scalar distance $\hat d_i$.

The probability head is supervised by a weighted binary cross-entropy,
the direction and distance heads by mean squared error for only the positive points. The loss is
\begin{equation}
\begin{aligned}
\mathcal{L}={}&-\frac{1}{N}\sum_{i=1}^{N}
\left[\alpha a_i\log\hat a_i+(1-a_i)\log(1-\hat a_i)\right]\\
&+\frac{\lambda_v}{\max(N_+,1)}\sum_{i=1}^{N}a_i
\left(\frac{\hat v_i}{\|\hat v_i\|_2}-v_i\right)^2\\
&+\frac{\lambda_d}{\max(N_+,1)}\sum_{i=1}^{N}a_i(\hat d_i-d_i)^2.
\end{aligned}
\label{eq:loss}
\end{equation}
Here $N_+=\sum_{i=1}^{N} a_i$ is the number of positive points,
and $\alpha$, $\lambda_v$, $\lambda_d$ are weighting coefficients.

\subsection{Stability Evaluation}
\begin{figure}[t]
	\centering
	\includegraphics[width=0.9\linewidth]{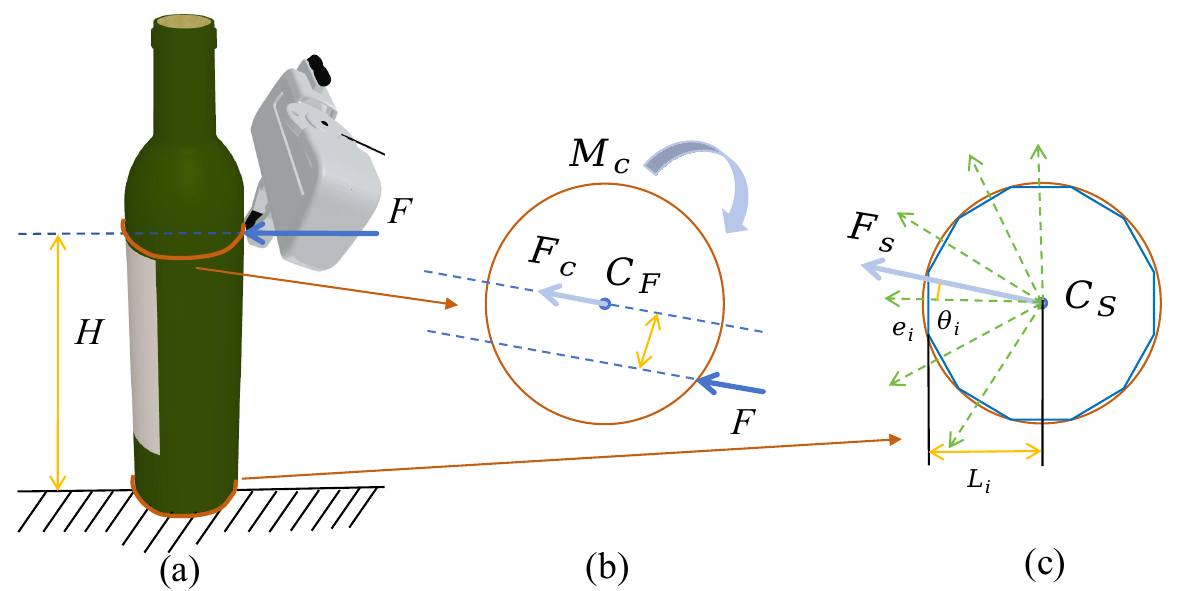}
	\caption{Quasi-static stability model. (a) A horizontal push is applied at height
    $H$. (b) On the horizontal contact plane, the force $F$ is decomposed about the projected center of mass $C_F$ into $F_c$ and $M_c$. (c) Support-plane, including the support polygon, the projection of $F_c$, the projected center of mass $C_S$ and candidate tipping edges. (In a more general setting, the blue convex polygon $\mathcal{H}$ is the support convex hull.); dashed arrows show outward normals.}
	\label{fig:stability-evaluation}
    \vspace{-0.7cm}
\end{figure}

We use a quasi-static model to evaluate the policy's contact candidates. For each candidate, $H$ is its contact height, and the push direction is horizontal. Let $m$ be the object mass, $g$ the gravitational acceleration, and $\mu$ the object-support friction coefficient.
The model compares the force required to initiate sliding with the force required to tip the object about an edge of its support polygon (Fig.~\ref{fig:stability-evaluation}).

At runtime, the support polygon is approximated by the convex hull of bottom-surface
points, and the center of mass is approximated by the point-cloud centroid. 
Let \(\mathcal{E}=\{e_i\}_{i=1}^{M}\) denote the set of edges of \(\mathcal{H}\).
For edge $e_i$, gravity acts at $C_S$, and the effective component of $F$ forms an angle $\theta_i$ with its outward normal.
Balancing the tipping and restoring moments yields the critical tipping force $F_{\mathrm{tip},i}=\frac{mgL_i}{H\cos\theta_i}$.
For fixed \(m\), \(g\), and \(H\), the edge most likely to induce toppling is therefore the one $i^\star=\arg\min_i\frac{L_i}{\cos\theta_i}$.
For $e_{i^*}$, $L$ is the perpendicular
distance from the projected center of mass to that edge, and $\theta$ is the angle
between the push direction and the edge's outward normal. The force required for
tipping is $mgL/(H\cos\theta)$, whereas that required for Coulomb sliding is $\mu mg$.
Their common factor $mg$ cancels, giving the sliding-before-tipping ratio
\begin{equation}
    \tau=\frac{L}{H\cos\theta}.
    \label{eq:stability-ratio}
\end{equation}
Under this model, sliding is predicted to start before tipping when $\mu<\tau$.
The ratio is dimensionless and unchanged by ideal uniform scaling of both $L$ and
$H$.
For evaluation, each candidate's learned contact probability is multiplied by
$\tanh(\beta\tau)$, and the candidate with the largest product is selected. Here
$\beta>0$ controls saturation. The ratio is a soft ranking cue, not a stability
constraint.

\section{Experiments}
\label{sec:new_experiments}

The experiments are designed to comprehensively evaluate the performance of our method on long-horizon, cluttered pushing tasks, including both simulated ones and real-world scenes. Besides, we also conduct ablation studies on the horizon $h$ as well as other proposed components.
Results demonstrate that: 
\begin{itemize}
    \item \textit{Our method can do reliable long-horizon pushing with limited training data, and directly generalize to the real world, including large-scale mobile pushing tasks;}
    \item \textit{Seeing forward with larger $h$ is not always beneficial due to inaccurate dynamics of pushing, but it helps significantly when there is \emph{high-risk} alternative paths;}
    \item The learned object-centric policy plays a crucial role for reliable performance, and failure-guided cost updates are also important for long-horizon pushing success.
\end{itemize}

\subsection{Experimental Setup}

\begin{figure}[t]
	\centering
	\includegraphics[width=0.95\linewidth]{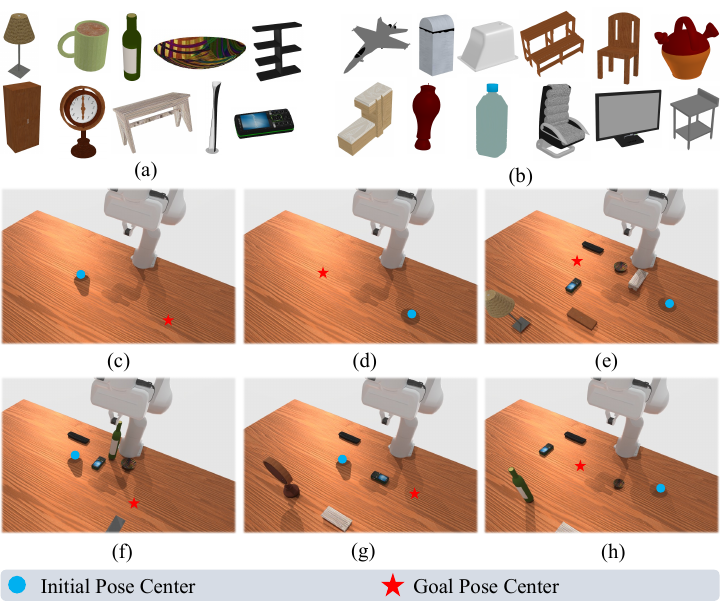}
    \vspace{-0.2cm}
	\caption{Simulation benchmark. Subfigures (a) and (b) show the tested
objects, with the former containing seen objects and the latter unseen
objects; Scenes~(c) and (d) are denoted as \textit{Empty}, Scenes~(e) and (f) as \textit{Cluttered}, and Scenes~(g) and (h) as \textit{Adversarial}.}
	\label{fig:simulation-bench-setting}
    \vspace{-0.5cm}
\end{figure}

We first evaluate the method with 22 ShapeNet objects \cite{shapenet2015}
in the six simulation scenes shown in Fig.~\ref{fig:simulation-bench-setting}, including 10 seen and 12 unseen objects.
In each trial, the robot pushes an object from an initial pose to a target pose. The initial and target positions are sampled within a circular region of diameter $5,\mathrm{cm}$, while their orientations are sampled uniformly at random. Each object is evaluated in 50 trials per setting.

For simulation, we reconstruct the object point cloud from three cameras and
obtain object masks using SAM2~\cite{ravi2024sam2}.
Object poses are estimated with the PCA-ICP pipeline.
To support diverse 3D geometries,
task success is defined as the square root of the Chamfer distance
between the current and target point clouds being below $1.5\,\mathrm{cm}$.
Each boxplot observation is an object-scene success rate from 50 trials, with a total of $22\times6=132$ observations, and 6600 long-horizon pushing trials.
For group-mean success rates, we compute 95\% percentile bootstrap intervals using 20,000 whole-object resamples within the 10 seen and 12 unseen objects,
retaining each object's scene and horizon records.
Our experiments are run on an RTX 2080 GPU and an Intel i9-14900KF CPU.

For real-world deployment, we use a dual-arm robotic system
with 4 objects in three real-world obstacle settings.
Each scene is set so that:
1) there exists a path (e.g., a line, a U-shape, or a S-shape) from the initial pose to the target pose;
2) obstacles and the target are visible.
Object poses are estimated from single RGB-D camera observations
using SAM3D~\cite{sam3dteam2025sam3d3dfyimages} and FoundationPose~\cite{10655554}.
For each object under each setting, we run 5 different trials.

\begin{figure*}[t]
\centering
\includegraphics[width=\textwidth]{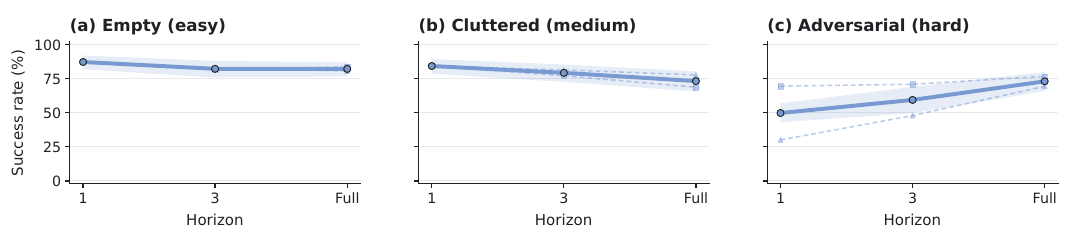}
\vspace{-0.7cm}
\caption{Horizon sensitivity in (a) empty (S1-2), (b) cluttered (S3-4), and (c) adversarial (S5-6) scenes. Solid lines show group means, dashed lines show individual scene means, and shading shows 95\% object-bootstrap confidence intervals. Each group mean uses 22 objects in two scenes.}
\label{fig:horizon-results}
\vspace{-0.2cm}
\end{figure*}

\subsection{Results with Different Horizons}
\label{sec:h_selection}

We evaluate $h=1$, $h=3$, and full-horizon planning in three kinds of scenes (Fig.~\ref{fig:horizon-results}). 
For Empty and Cluttered scenes, $h=1$ achieves the highest
mean success of 87.23\% and 84.27\%.
By contrast, full-horizon planning yields 82.14\%
and 73.23\%, respectively.
In these tasks, a longer planning horizon may accumulate prediction errors during forward search, and hence deviate from the underlying true optimal path.
As a result, failures here are mainly due to planning, in which it filters out some physically feasible paths and finds no solutions.
For the Adversarial scenes, which include an apparently feasible object path whose later robot motions are infeasible, success rises from 49.82\% at
$h=1$ to 73.05\% with full-horizon planning, which highlights the significance of long-horizon planning.

\begin{figure*}[t]
\centering
\includegraphics[width=.333\textwidth]{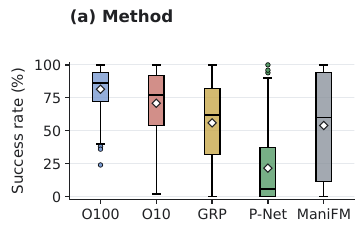}%
\includegraphics[width=.333\textwidth]{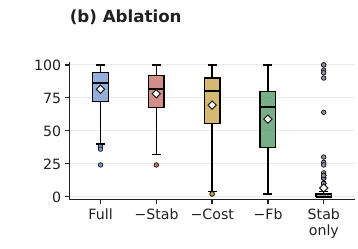}%
\includegraphics[width=.333\textwidth]{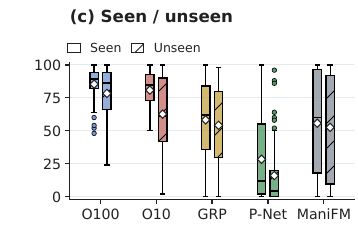}
\vspace{-0.7cm}
\caption{Success-rate distributions for (a) methods, (b) ablations, and (c) seen/unseen objects. Each box contains 132 object-scene rates in (a,b), or 60/72 in (c), with 50 trials per rate. Boxes span the interquartile range; black lines mark medians, diamonds mark means, whiskers use 1.5 times the interquartile range, and circles mark outliers. O100/O10 denote Ours-100/Ours-10; Full denotes Ours-100.}
\label{fig:comparison-results}
\vspace{-0.5cm}
\end{figure*}

\subsection{Results of Different Methods}
We compare to 3 baselines, including: \textit{GRP}~\cite{Li2018PushNet} (Greedy Reactive Policy with stability scoring),
which selects goal-conditioned contacts heuristically and evaluates them with our stability model;
\textit{P-Net} (Push-Net)~\cite{Li2018PushNet}, adapted with multi-view
reconstruction, rotation augmentation, and object-centric inputs; and
\textit{ManiFM} (ManiFoundation)~\cite{10801782}, adapted with motion-feature
extraction, sparse-contact-aware training, and robot-policy decoupling.
Besides, we include an additional object-centric baseline, Ours-10, which is only trained with 10 objects.
By comparison, Ours-100 is the default model, which uses 100 training objects with the same total number of training examples.
Since the high-level planner setting $h$ severely affects the end-to-end testing performance, for fair comparisons, we use $h=1$ for the Empty and Cluttered scenes, and a full-horizon planner for the Adversarial scenes for all baselines and our methods.

\begin{table}[t]
\centering
\caption{Mean success rates (\%). Each method uses 6,600 trials}
\label{tab:success-rate-different-method}
\vspace{-7pt}
\small
\setlength{\tabcolsep}{10pt}
\begin{tabular}{@{}lrrr@{}}
\toprule
Method & All objects & Seen objects & Unseen objects\\
\midrule
Ours-100 & \textbf{81.52} & \textbf{85.37} & \textbf{78.31} \\
Ours-10 & 70.89 & 80.83 & 62.61 \\
GRP & 55.97 & 58.23 & 54.08 \\
P-Net & 21.59 & 28.33 & 15.97 \\
ManiFM & 53.95 & 55.70 & 52.50 \\
\bottomrule
\end{tabular}
\vspace{-10pt}
\end{table}

Table~\ref{tab:success-rate-different-method} reports mean success rates,
while Fig.~\ref{fig:comparison-results}(a,c) shows their distribution across
objects and scenes. Ours-100 reaches 81.52\% overall and 78.31\% on unseen
objects. Increasing the training objects improves seen success by 4.54\% and unseen success by 15.70\%.
The boxplots also expose variation hidden by scene averages: P-Net has a
6\% median and 34 completely failed object-scene combinations; ManiFM's
interquartile range is 11.5-94\%, with 13 completely failed combinations.
Ours-100 has no zero-success combination and an interquartile range of
72-94\%. These comparisons support transfer within the evaluated object
set.

\begin{table}[t]
    \centering
    \caption{Comparison with CEM-MPC.}
    \label{tab:main-mpc-comparison}
    \vspace{-7pt}
    \setlength{\tabcolsep}{6pt}
    \renewcommand{\arraystretch}{0.92}
 \small
        \begin{tabular}{@{}llcccc@{}}
            \toprule
            Object & Method & Position & Orientation & Time & Falls \\
            type   &        & (cm) & (rad) & (s) &       \\
            \midrule
            Seen & Ours    & \textbf{1.31} & \textbf{0.466} & \textbf{0.026} & \textbf{2} \\
            Seen & CEM-MPC & 3.41 & 0.608 & 81.8 & 14 \\
            \midrule
            Unseen & Ours    & \textbf{1.82} & \textbf{0.463} & \textbf{0.027} & \textbf{4} \\
            Unseen & CEM-MPC & 3.44 & 0.557 & 71.8 & 6 \\
            \bottomrule
        \end{tabular}
        \vspace{-0.2cm}
 \end{table}

We also compare our method with a model-based approach,
\textit{CEM-MPC}, under a short-horizon pushing setting. It uses MUJOCO simulation for rollouts and the cross-entropy method for optimization~\cite{11128433}.
In this setting, the robot is allowed at most two pushing actions to relocate the object to a near target pose.
The results are shown in Table~\ref{tab:main-mpc-comparison}.
We report the final pose error between the object and the goal, whether the object tips over,
and the inference time for a single push action.
Within this two-push setting, our method has lower reported position and orientation
errors, fewer falls, and lower per-action inference time than CEM-MPC.

\begin{figure*}[ht]
	\centering
	\includegraphics[width=\textwidth]{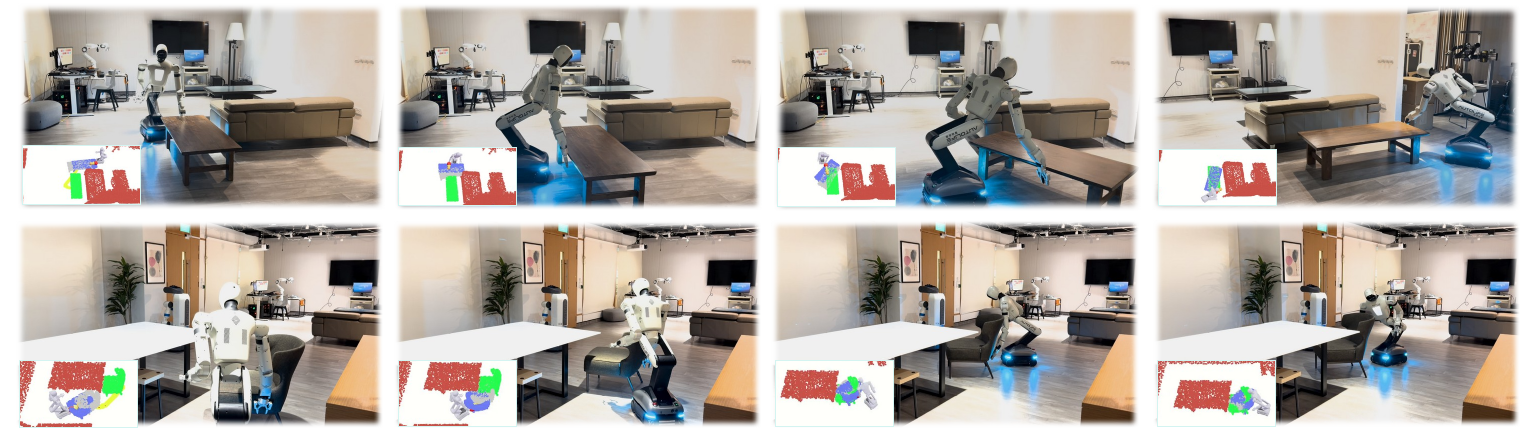}
	\caption{Real-robot large-scale object relocation with a mobile manipulator. Top row: a table. Bottom row: a chair.}
	\label{fig:mobile-setting}
    \vspace{-0.6cm}
\end{figure*}

\begin{table}[t]
 \centering
	\caption{Tabletop successes and mean success rate (\%).}
	\label{tab:main-real-world-success}
    \vspace{-7pt}
	\setlength{\tabcolsep}{8pt}
	\renewcommand{\arraystretch}{0.9}
	\small
	\begin{tabular}{lcccc}
		\toprule
		Object & Scene 1 & Scene 2 & Scene 3 & Average \\
		\midrule
		Packing tape & 3/5 & 4/5 & 4/5 & 73.3\\
		Hammer       & 2/5 & 2/5 & 3/5 & 46.7\\
		Bottle       & 4/5 & 3/5 & 3/5 & 66.7\\
		Chair        & 4/5 & 4/5 & 4/5 & 80.0\\
		\bottomrule
	\end{tabular}
    \vspace{-0.6cm}
 \end{table}

\begin{figure}[t]
	\centering
	\includegraphics[width=\linewidth]{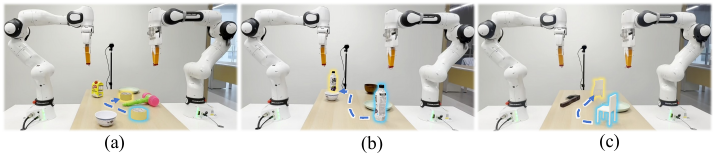}
    \vspace{-0.7cm}
	\caption{Dual-arm robot scenes. The object marked in blue indicates the initial position, while the object inside the yellow indicates the target pose. (a), (b), and (c) show the scenarios with 4, 3, and 2 obstacles, respectively.}
	\label{fig:dual-arm-setting}
    \vspace{-0.5cm}
\end{figure}

\begin{figure}[t]
\centering
\includegraphics[width=0.92\linewidth]{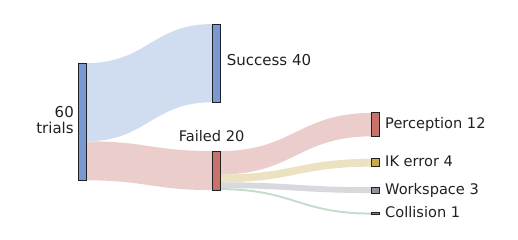}
\vspace{-0.6cm}
\caption{Outcomes of 60 tabletop trials. Ribbon widths are proportional to counts; the 20 failures are partitioned by recorded cause.}
\label{fig:real-failures}
\vspace{-0.7cm}
\end{figure}

\subsection{Ablation Study}

We demonstrate comparisons among four variants in Figure~\ref{fig:comparison-results}(b):
\textit{$-$Stab}, without stability scoring;
\textit{$-$Cost}, without local path-cost updates after failed robot motion planning;
\textit{$-$Fb}, without future feasibility check and feedback; and
\textit{Stab only}, selecting contacts using only the stability score, as a heuristic ablation.
The mean (diamonds) are 81.52\%, 78.06\%, 69.38\%, 58.77\%, and
6.55\%, respectively. 
Notably, \textit{Stab only} has a 0\% median with 81/132 zero-success pairs, indicating that only stability scores alone are insufficient for reliable goal-directed
pushing in these tasks.
Particularly, the cost-update difference is concentrated in the adversarial scenes.
This is because the planner does not inherently favor routes near feasible regions, requiring feedback for replanning.
Stability scoring brings marginal performance gain, but leads to better performance for some geometrically complex objects.
For example, on the eighth unseen object, a vase (Fig.~\ref{fig:simulation-bench-setting}(b)), the overall success rate drops from $93\%$ to $50\%$ without the stability evaluation (300 trials in 6 scenes).

\subsection{Real-world Deployment}
\label{sec:real-world}

We conduct real-world experiments on two Franka FR3 arms mounted on opposite sides of a shared tabletop and an AutoLife mobile manipulator.
The tabletop setup uses an Intel RealSense L515 RGB-D camera for perception and a VAMP \cite{vamp_2024} planner for collision-aware motion planning, with one arm executing each push.
AutoLife leverages the object-level planner and learned pushing policy, integrating base navigation with whole-body IK and collision-aware planning over the leg, waist, and arm joints.
Both re-estimate the object pose and replan after each push.
Across the tabletop experiments (Fig.~\ref{fig:dual-arm-setting}), the system succeeds in 40/60 trials (66.7\%; Table~\ref{tab:main-real-world-success}).
We also include a full failure analysis for the real-robot deployment in Fig.~\ref{fig:real-failures}) for a deeper understanding of the system-level bottleneck.
Among all failures, perception accounts for more than half, 12/20 failures.
The remaining comprises mainly four types of failures including infeasible IK (IK error), object pushed out of workspace (Workspace), and collision failure (Collision).
Perception is therefore the main recorded limitation, revealing that geometry or pose errors can affect planning, push action prediction, and feasibility checking, motivating better reconstruction and tracking alongside pushing components.

Finally, we deploy our method on the AutoLife mobile robot to relocate a chair and a coffee table in household environments (Fig.~\ref{fig:first-figure}). 
The complete relocation process for both objects is shown in Fig.~\ref{fig:mobile-setting}.

\section{Conclusion}
\label{sec:conclusion}
We presented a long-horizon pushing system that integrates object-level path search, object-centric push policy learning, quasi-static stability ranking, and robot-motion planning in a hierarchical planning framework. Performance in both simulation and real-world tabletop experiments supports generalization to unseen object geometries and shows benefits from the feedback and stability modules.
The mobile-manipulator demonstrations show that our model can directly generalize to large-scale objects, without retraining or data collection.
The main system-level bottleneck lies in the object-centric perception pipeline, which relies on stable object pose estimation and tracking.
Future work can consider enhancing perception modules \cite{anand2026reconstruction} or absorbing perception noise into data-driven policies.
Besides, for non-prehensile manipulation like pushing, physical parameters like the center of mass, friction, and their uncertainties are critical for robustness.
Future work includes estimating these parameters with uncertainty-aware models and integrating our pushing policy into longer-horizon tasks that require diverse skill primitives \cite{zandonati2025rational, huang2026kinder, shen2026tiptop}.

\section*{Acknowledgment}

We used vibe coding tools including Codex and Claude Code to help implement parts of the algorithm and analyze bugs. Besides, we used AI tools to design visuals for Figure \ref{fig:horizon-results}, \ref{fig:comparison-results}, and \ref{fig:real-failures}, and to polish the text and grammar.




\bibliographystyle{IEEEtran}
\bibliography{references}

\newcommand{\supv}[3]{\(\text{#1}^{\text{+#2}}_{\text{-#3}}\)}
\newcommand{\supb}[3]{\(\textbf{\text{#1}}^{\textbf{\text{+#2}}}_{\textbf{\text{-#3}}}\)}

\section{Appendix}
\label{sec:appendix}

This appendix provides supplementary material organized as follows:
stability evaluation, receding-horizon planning, contact-to-robot-motion
mapping, data generation, and the pushing termination criterion
(Sections~\ref{app:stability-details}--\ref{app:termination}); real-world
deployment details for the dual-arm tabletop system and the AutoLife mobile manipulator (Section~\ref{app:real-world-deployment}); baseline implementation details (Section~\ref{app:baselines}); and additional results on horizon selection, method comparisons, ablation studies, and the impact of pose estimation (Sections~\ref{app:horizons}--\ref{app:pose-estimation}).

\subsection{Stability Evaluation Details}
\label{app:stability-details}

Write an edge of \(\mathcal H\) as
\(n_i^Tx=b_i\), where \(n_i\) is its unit outward normal. Its moment arm is
\(L_i=b_i-n_i^TC_S>0\). Let
\(\bar v_{xy}=v_{xy}/\lVert v_{xy}\rVert_2\) be the horizontal projection of
the predicted push direction and
\(\cos\theta_i=n_i^T\bar v_{xy}\). Only edges with
\(\cos\theta_i>0\) can be tipping axes in the direction of the applied push.

For a horizontal force \(F\) applied at height \(H\), the overturning and
restoring moments about edge \(i\) are
\begin{equation}
M_{F,i}=FH\cos\theta_i,
\qquad
M_{G,i}=mgL_i.
\label{eq:appendix-moments}
\end{equation}

Therefore,
\begin{equation}
F_{\mathrm{tip},i}=\frac{mgL_i}{H\cos\theta_i},
\qquad
i^\star=\arg\min_{i:\cos\theta_i>0}
\frac{L_i}{\cos\theta_i}.
\label{eq:appendix-tipping-edge}
\end{equation}

The Coulomb sliding threshold is \(F_{\mathrm{slide}}=\mu mg\). For the most
critical edge, \(F_{\mathrm{slide}}<F_{\mathrm{tip}}\) precisely when
\begin{equation}
\mu<\tau,
\qquad
\tau=\min_{i:\cos\theta_i>0}\frac{L_i}{H\cos\theta_i},
\label{eq:appendix-tau-complete}
\end{equation}

Mass and gravity cancel and a common unknown friction coefficient also need not be estimated to rank
contacts on the same object and support. Moreover, \(L_i\) and \(H\) scale
together, so \(\tau\) is dimensionless and invariant to uniform object scale.
A large \(\tau\) favors sliding before tipping, while a small value indicates
a small stability margin.
At execution time, all points with $\hat{a}_i>0.5$ or the top-$10$ highest-probability points are used. 
Contacts in the bottom are assigned zero stability score 
to avoid numerical degeneracy near the support surface.

\subsection{Receding-Horizon Planning}
\label{app:receding-horizon}

The high-level and low-level planners are coupled in a receding-horizon loop.
BIT\(^*\) first returns a collision-free object-pose path
\(\Gamma=X_{k:K}\).
The path is then interpolated into single-push targets.
During forward motion planning and feasibility analysis, 
we assume that the object can reach each single-push target as intended. 
Each such step is termed a \textit{virtual push primitive}, 
and the horizon $h$ denotes the number of primitives evaluated ahead.
At every virtual step, the policy predicts contacts from the virtually
transformed cloud, the stability model evaluates the candidates, and the robot
motion planner verifies the complete push primitive. 
Consequently, this look-ahead tests kinematic reachability and collision feasibility. 
Only the first verified primitive is executed, after which the object is perceived again.
Algorithms~\ref{alg:rh-planning} and~\ref{alg:prefix-validation} provide the complete procedure.

\setlength{\textfloatsep}{0.7\baselineskip  plus 0.1\baselineskip minus 0.1\baselineskip}

\begin{algorithm}[!t]
\caption{Failure-aware receding-horizon pushing}
\label{alg:rh-planning}
\Input{goal \(X_g\), horizon \(h\), margin candidates $\mathcal{B}$, maximum push count
\(N_{\mathrm{p}}\), and validation-attempt limit \(N_{\mathrm{v}}\)}
\State{failed-pose memory \(\mathcal F_{\mathrm p}\), failed-transition
memory \(\mathcal F_{\mathrm t}\), and cached object path \(\Gamma\)}
Initialize \(k\leftarrow0\),
\(\mathcal F_{\mathrm p}\leftarrow\varnothing\),
\(\mathcal F_{\mathrm t}\leftarrow\varnothing\), and
\(\Gamma\leftarrow\varnothing\)\;
\While{\(k<N_{\mathrm{p}}\)}{
  Observe \(P_k\), estimate \(Y_k\), and read the robot configuration
  \(q_k\)\;
  Set keyframe ee-poses of a push primitive \(a^{\star}\leftarrow\varnothing\) \;
  \For{\(r\leftarrow1\) \KwTo \(N_{\mathrm{v}}\)}{
    \If{\(\Gamma=\varnothing\)}{
      Use \(\mathcal F_{\mathrm p},\mathcal F_{\mathrm t}\) to update BIT\(^*\) palnner\; 
      Pop margins from $\mathcal{B}$ in descending order and plan path $X_{k:K}$\;
      Set \(\Gamma \leftarrow X_{k:K}\)\;
    }
    \If{$\Gamma= \varnothing$}{
      \textbf{continue}\;
    } 
    Run \mbox{\textsc{ValidateHorizon}}\((\Gamma,P_k,X_k,Y_k,q_k,h)\)\;
    \If{validation succeeded}{
      Update \(a^{\star}\) from validation result and \textbf{break}\;
    }
    Extract \(\bar X\) and \(\bar e\) from validation result\;
    Add \(\bar X,\bar e\) into
    \(\mathcal F_{\mathrm p},\mathcal F_{\mathrm t}\), update the cost distribution, and set
    \(\Gamma\leftarrow\varnothing\)\;
  }
  \KwSty{end}\;
  \If{\(a^{\star}=\varnothing\)}{
    \textbf{return} failure\;
  }
  Execute \(a^{\star}\)\;
  \If{the object reaches the goal}{
    \textbf{return} success\;
  }
  \If{the object has fallen}{
    \textbf{return} failure\;
  }
  Set \(k\leftarrow k+1\)\;
}
\KwSty{end}\;
\textbf{return} failure (push budget exhausted)\;
\end{algorithm}

\begin{algorithm}[!t]
\caption{\textsc{ValidateHorizon}}
\label{alg:prefix-validation}
\Input{path \(\Gamma\), horizon \(h\) and observed state
\((P_k,X_k,Y_k,q_k)\)}
\Output{validation result}
Initialize
\((\widetilde P,\widetilde X,\widetilde Y,\widetilde q)
\leftarrow(P_k,X_k,Y_k,q_k)\)\;
\For{\(s \leftarrow 0 \) \KwTo \(h-1\)}{
  \If{the end of \(\Gamma\) has been reached}{
    \textbf{return} feasible, first virtual push primitive\;
  }
  Interpolate $\Gamma$ to obtain the
  one-push target \(\widetilde X^+\)\;
  Construct the desired canonical transform \(T\) and canonical cloud
  \(\widetilde P^{(\phi)}\) using Eq.~\eqref{eq:canonical-inputs}\;
  Predict dense \(\{(\hat a_i,\hat v_i,\hat d_i)\}\), compute \(\tau_i\), and
  select \(u_s=\arg\max_i(\hat a_i * \tanh(\beta\tau_i))\)\;
  Map \(u_s\) to end-effector poses, then motion plan Sec.~\ref{app:contact-to-ee}.\;
  \If{motion plan fails}{
    \textbf{return} infeasible, failed pose $\widetilde X$ and \(\widetilde X^+\) \;
  }
  Update \(\widetilde q\) to the planned terminal robot configuration\;
  Transform \(\widetilde P\) from
  \(\widetilde X\) to \(\widetilde X^+\), update \(\widetilde Y\)
  and set \(\widetilde X\leftarrow\widetilde X^+\)\;
}
\KwSty{end}\;
\textbf{return} feasible, first virtual push primitive\;
\end{algorithm}

We use two Gaussian kernels to update the edge-cost distribution of the path planner (Eq.~\ref{eq:cost-update}). 
Here, $d_t$ denotes the distance from a point on a edge $e$ to its nearest point on $\bar{e}$,
$d_\theta$ denotes the directional difference between edges $e$ and $\bar{e}$.
$5$ is a hyperparameter specifying the peak cost density. 
As it approaches zero, the additional penalty on edges near failed regions vanishes; 
larger values impose stronger penalties.
During path planning, the cost of edge $e$ is computed according to Eq.~\ref{eq:appendix-integrated-cost}.
Here, $D(e)$ denotes the $SE(2)$ distance, while the integral term represents the additional cost 
induced by failure feedback.
This cost-based feedback penalizes failed regions without declaring them geometrically forbidden, 
allowing the planner to reuse them if they later become the only feasible option.
\begin{equation}
\begin{aligned}
\bigl[1+C(e)\bigr]D(e)
&=D(e)+\int_e
\left[c_{\mathrm p}(x)+c_{\mathrm t}(x,e)\right]\mathrm d\ell
\end{aligned}
\label{eq:appendix-integrated-cost}
\end{equation}

BIT\(^*\) uses \(20\) batches of \(200\) samples. 
The cost integral is evaluated by the trapezoidal rule at the \(1\,\mathrm{cm}\) edge-checking resolution.
Edges are collision checked at \(1\,\mathrm{cm}\) resolution and the returned path is processed by
cost-aware shortcutting, which accepts a shortcut only if it does not increase
the path cost. Object clearance margins of \(5\), \(2\), and \(1\,\mathrm{cm}\)
are tried in sequence, so the planner prefers generous clearance.
The planner stores at most five failed poses and five failed transitions. 
Failed poses and transitions that are too close to existing entries are discarded 
to avoid excessive cost accumulation in local regions.

\subsection{From Contact to Robot Motion}
\label{app:contact-to-ee}

Let \(u=(p,v,d)\) be the selected canonical action and let \(\ell\) be the
current normalization length from Sec.~\ref{subsec:policy}. Its world-frame contact,
direction, and distance are
\begin{equation}
c=Q_kp\ell+o_k,
\qquad
v^w=\frac{Q_kv}{\lVert Q_kv\rVert_2},
\qquad
d^w=d\ell.
\label{eq:appendix-world-action}
\end{equation}

The contact \(c\) is not itself the initial TCP position. With approach gap
\(\delta_a=0.05\,\mathrm m\) and endpoint shortening
\(\delta_e=0.01\,\mathrm m\), the TCP translations are
\begin{equation}
x_{\mathrm s}=c-\delta_a v^w,
\qquad
x_{\mathrm e}=c+(d^w-\delta_e) v^w.
\label{eq:appendix-ee-translations}
\end{equation}

The yaw angle of the end-effector TCP during pushing is defined as $\psi=\operatorname{atan2}(v_y^w,v_x^w)$, 
where $v_x^w$ and $v_y^w$ are the $x$- and $y$-components of $v^w$, respectively.
Start and end use the same TCP orientation
\begin{equation}
R_E=R_z(\psi)R_y(-\gamma)R_x(\pi),
\label{eq:appendix-ee-orientation}
\end{equation}
where $\gamma$ denotes the desired TCP pitch for object contact.

We decompose a complete push primitive into three phases: approach, push, and
retract. The push phase moves the TCP from $x_{\mathrm s}$ to
$x_{\mathrm e}$ using the screw-motion planner provided by ManiSkill3. During
the approach phase, the TCP first moves to a pose above $x_{\mathrm s}$ and
then descends to $x_{\mathrm s}$, with both motions planned by RRTConnect.
During the retract phase, the TCP first returns from $x_{\mathrm e}$ to
$x_{\mathrm s}$ using the screw-motion planner and then moves back to the pose
above $x_{\mathrm s}$ using RRTConnect.

\subsection{Data Generation Supplement}
\label{app:data_generation}

We construct the 100-object trainset from ShapeNetCore.v2~\cite{shapenet2015}
using a geometry-diversity procedure. 
We recursively enumerate
all files and retain a mesh only without a skipped non-triangular primitive. 
This produces 14,135 valid candidate meshes from the 55 categories. 
For every candidate mesh, we uniformly sample $2048$ surface points.
Then we remove its sample mean and normalize its maximum radius.
Thus, object translation and absolute scale do not affect the diversity
criterion. Each normalized $2048\times3$ array is flattened and projected to
a 50-dimensional PCA embedding. We compute Euclidean distances in this
embedding and apply greedy max--min farthest-point sampling. Starting from a
random candidate and a selected set $\mathcal S$, each new object is chosen as
\begin{equation}
j^\star=\arg\max_{j\notin\mathcal S}
\min_{i\in\mathcal S} d(i,j).
\label{eq:appendix-mesh-fps}
\end{equation}
Using the embedding distance for $d$ gives a coarse shortlist of 2,000 meshes.

\begin{figure*}[ht]
	\centering
    \includegraphics[width=\textwidth]{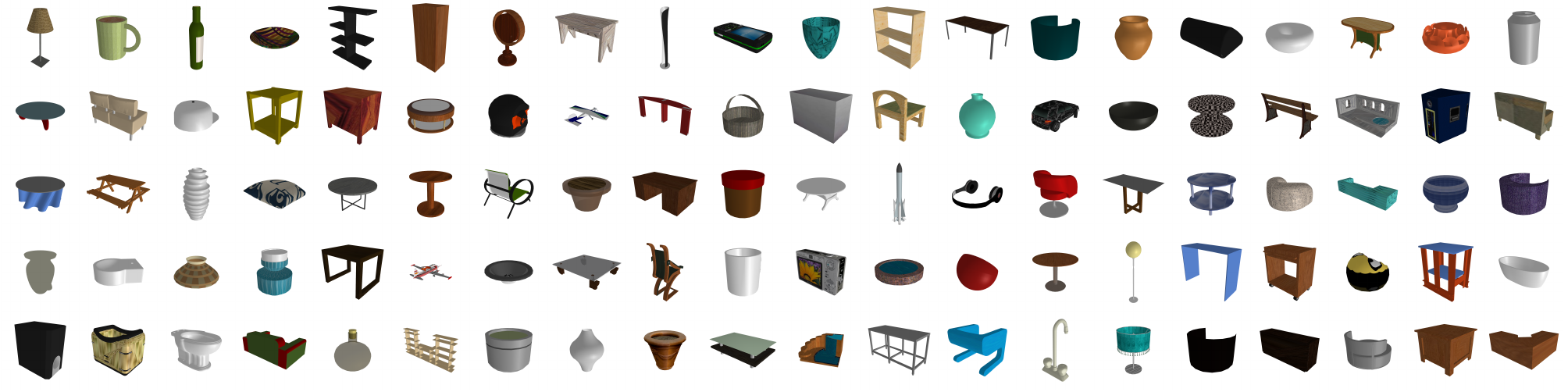}
    \vspace{-0.3cm}
	\caption{The 100 trainset objects.}
	\label{fig:training-objects}
    \vspace{-0.6cm}
\end{figure*}

We then refine diversity using the sampled geometry. For each pair in the
shortlist, one normalized cloud is aligned to the other by point-to-point ICP,
initialized with the identity transform and using a correspondence threshold
of 0.05 in normalized coordinates. If $\widehat P_i^{(j)}$ denotes cloud $i$
after alignment to cloud $j$, the implementation uses the one-way
nearest-neighbor score
\begin{equation}
d_{\mathrm{NN}}(i,j)=
\sum_{p\in\widehat P_i^{(j)}}\min_{q\in\widetilde P_j}
\lVert p-q\rVert_2^2.
\label{eq:appendix-mesh-distance}
\end{equation}
The computed value is stored for that pair, and a second
max--min sampling pass with this distance retains the final 100 models.
Finally, models incompatible with the simulation environment or unsuitable for data collection are removed 
and replaced with additional objects, yielding the final training set shown in Fig.~\ref{fig:training-objects}.

Moreover, the training object scale range is defined independently of the test-time scales, 
without being manually adjusted to match them.

\subsection{Terminal Condition for Pushing}
\label{app:termination}

In the real world, the ground-truth pose of an object cannot be directly obtained
and must instead rely on pose estimation algorithms.
However, existing methods often struggle to accurately estimate the orientation of certain symmetric objects,
such as cylindrical objects.
We therefore use the square-root Chamfer distance $d(P_k, P_g)$ between the current 
and goal object point clouds to measure task progress, where $P_g$ is determined by the initial object cloud, 
the initial pose, and the goal pose.
The task is terminated when $d(P_k, P_g) < 0.015m$.

\subsection{Real World Deployment}
\label{app:real-world-deployment}

\subsubsection{Dual-Arm Tabletop System}
\label{app:dual-arm}

We conduct the tabletop real-world experiments using two Franka FR3 arms mounted 
on opposite sides of a shared tabletop workspace, enabling pushes from either side of the object. 
The arms share the same calibrated workspace, and all robot motions are planned in the robot world frame. 
Each push is executed by one arm at a time; when switching arms, 
the previously active arm is first returned to its home configuration before the other arm approaches the object.

The scene is observed by a single Intel RealSense L515 RGB-D camera mounted to view the tabletop workspace. 
At the beginning of each trial, the user specifies the target object and the obstacles in the tabletop scene. 
After this specification, the perception module segments the scene, reconstructs the object and obstacle point clouds, and initializes object tracking. 
During execution, the object pose is updated after each push, while camera frames collected during the pushing stroke are used to improve tracking continuity.

For robot motion planning, we use a bimanual planner built on VAMP and OMPL. 
The planner supports active subgroup planning, which allows each motion to be planned for only the currently active arm while keeping the full dual-arm system and workspace collision model in consideration. 
For free-space arm motions, the planner considers the static workspace geometry, tabletop obstacles, and the current object point cloud. 
During the pushing stroke and retraction, the object point cloud is removed from the collision cloud so that the intended contact configuration is not rejected as collision. 
If the motion plan for the preferred arm is infeasible, the system attempts the motion with the other arm.

The planned trajectories are executed on the Franka arms through the default Franka-supported position-control interface. 
The pushing stroke is executed with a lower velocity scale than the approach and retraction motions to improve contact stability. 
A trial ends when the object satisfies the goal tolerance or when the maximum number of allowed pushes is reached. 
In the real world, the object mesh and ground-truth pose cannot be directly obtained. 
Therefore, the results of the pose estimation algorithm are used to determine whether the trajectory succeeds.

\subsubsection{AutoLife Mobile Manipulator}
\label{app:autolife}

For the mobile manipulation experiment, we deploy the proposed pushing framework on the AutoLife mobile manipulator. 
The object-level planner and learned pushing policy remain unchanged: at each iteration, 
the system estimates the object pose, plans the next object subgoal, 
and predicts a contact point, pushing direction, and pushing distance. 
The predicted push is then converted into a mobile whole-body action.

Given the contact and push direction, we sample feasible mobile-base poses behind the contact point 
and rank them by distance to the current base pose. 
For each candidate, the system selects the pushing arm according to the relative side of the base, 
solves whole-body IK for three key poses, namely pre-contact, contact, and push-end, 
and verifies the motion with collision-aware planning over the leg, waist, and arm joints. 
During the pushing stroke, an additional line-tracking cost keeps the end-effector moving along 
the intended push direction. The selected base pose is reached using the navigation stack, 
and the planned leg-waist-arm trajectory is executed on hardware. After each push, 
the object pose is re-estimated and the planner replans in closed loop until the object reaches the goal.

\begin{table*}[t]
    \centering
    \caption{Scene-wise success rates (\%, left) and mean push steps over
    successful trials (right) with different horizons.}
    \label{tab:app-horizon-combined}
    \vspace{-7pt}
    \footnotesize
    \setlength{\tabcolsep}{4pt}
    \renewcommand{\arraystretch}{1.08}
    \begin{tabular}{@{}l*{6}{c}@{\hspace{15pt}}*{6}{c}@{}}
        \toprule
        & \multicolumn{6}{c}{Success rate (\%) $\uparrow$}
        & \multicolumn{6}{c}{Push steps $\downarrow$}\\
        \cmidrule(r{15pt}){2-7}\cmidrule(l{0pt}){8-13}
        Horizon & S1 & S2 & S3 & S4 & S5 & S6
                & S1 & S2 & S3 & S4 & S5 & S6\\
        \midrule
        \multicolumn{13}{@{}l}{\textit{All Objects}}\\
        \addlinespace[4pt]
        1 & \supb{87.4}{5.2}{6.0} & \supb{87.1}{4.8}{5.5} & \supb{84.4}{6.1}{7.3} & \supb{84.2}{4.7}{5.4} & \supv{69.5}{6.1}{6.5} & \supv{30.1}{9.7}{8.7}
          & \supb{10.0}{0.3}{0.3} & \supb{10.0}{0.3}{0.3} & \supb{13.3}{0.5}{0.4} & \supb{9.8}{0.3}{0.3} & \supv{11.7}{0.4}{0.3} & \supb{11.9}{0.6}{0.6}\\
        \addlinespace[4pt]
        3 & \supv{81.5}{6.2}{6.5} & \supv{82.9}{6.0}{6.9} & \supv{77.1}{8.1}{9.0} & \supv{81.5}{5.1}{5.6} & \supv{70.9}{7.3}{7.9} & \supv{47.9}{12.6}{12.5}
          & \supv{10.2}{0.4}{0.3} & \supv{10.4}{0.4}{0.3} & \supv{13.5}{0.4}{0.4} & \supv{10.0}{0.3}{0.3} & \supv{11.8}{0.5}{0.5} & \supv{13.9}{0.5}{0.5}\\
        \addlinespace[4pt]
        Full & \supv{83.6}{4.9}{5.1} & \supv{80.6}{6.0}{6.3} & \supv{68.7}{9.3}{10.2} & \supv{77.7}{5.9}{6.6} & \supb{76.7}{6.5}{7.1} & \supb{69.4}{8.5}{9.2}
             & \supv{10.7}{0.4}{0.4} & \supv{10.9}{0.4}{0.4} & \supv{13.9}{0.4}{0.4} & \supv{9.9}{0.3}{0.3} & \supb{11.5}{0.4}{0.4} & \supv{12.4}{0.5}{0.5}\\
        \midrule
        \multicolumn{13}{@{}l}{\textit{Seen Objects}}\\
        \addlinespace[4pt]
        1 & \supb{89.4}{6.2}{7.0} & \supb{89.4}{5.0}{6.2} & \supb{86.0}{6.8}{7.6} & \supb{88.8}{4.6}{5.6} & \supv{70.4}{9.2}{10.4} & \supv{27.8}{14.8}{12.2}
          & \supb{10.1}{0.3}{0.3} & \supb{10.1}{0.4}{0.3} & \supb{12.8}{0.4}{0.4} & \supb{9.6}{0.3}{0.3} & \supv{11.6}{0.2}{0.3} & \supb{11.2}{1.0}{0.9}\\
        \addlinespace[4pt]
        3 & \supv{85.0}{7.4}{7.6} & \supv{86.0}{7.2}{8.0} & \supv{84.0}{8.0}{8.6} & \supv{85.8}{5.4}{5.8} & \supv{73.8}{10.4}{11.0} & \supv{46.8}{18.8}{18.8}
          & \supv{10.2}{0.3}{0.3} & \supv{10.4}{0.4}{0.3} & \supv{13.2}{0.3}{0.3} & \supv{9.7}{0.3}{0.3} & \supv{11.8}{0.4}{0.5} & \supv{14.2}{0.5}{0.5}\\
        \addlinespace[4pt]
        Full & \supv{83.6}{7.6}{8.0} & \supv{83.6}{7.2}{7.6} & \supv{75.0}{11.6}{13.0} & \supv{85.2}{4.8}{5.2} & \supb{86.0}{4.4}{5.0} & \supb{72.6}{10.0}{10.4}
             & \supv{10.6}{0.3}{0.3} & \supv{10.7}{0.5}{0.4} & \supv{13.7}{0.4}{0.4} & \supv{9.8}{0.3}{0.3} & \supb{11.3}{0.4}{0.4} & \supv{12.1}{0.3}{0.3}\\
        \midrule
        \multicolumn{13}{@{}l}{\textit{Unseen Objects}}\\
        \addlinespace[4pt]
        1 & \supb{85.7}{7.8}{9.2} & \supb{85.2}{7.8}{8.7} & \supb{83.0}{9.7}{11.7} & \supb{80.3}{7.2}{8.0} & \supv{68.8}{8.2}{8.5} & \supv{32.0}{13.7}{12.0}
          & \supb{9.9}{0.5}{0.4} & \supb{10.0}{0.5}{0.4} & \supb{13.7}{0.7}{0.7} & \supb{9.9}{0.4}{0.4} & \supv{11.8}{0.6}{0.5} & \supb{12.5}{0.6}{0.6}\\
        \addlinespace[4pt]
        3 & \supv{78.7}{9.3}{10.0} & \supv{80.3}{9.2}{10.8} & \supv{71.3}{12.4}{14.0} & \supv{77.8}{8.2}{8.5} & \supv{68.5}{10.0}{10.8} & \supv{48.8}{17.0}{17.3}
          & \supv{10.3}{0.6}{0.5} & \supv{10.4}{0.7}{0.6} & \supv{13.8}{0.6}{0.6} & \supv{10.1}{0.6}{0.5} & \supv{11.8}{0.9}{0.7} & \supv{13.6}{0.8}{0.8}\\
        \addlinespace[4pt]
        Full & \supv{83.7}{6.0}{7.0} & \supv{78.2}{8.6}{10.0} & \supv{63.5}{13.7}{14.8} & \supv{71.5}{9.2}{10.0} & \supb{69.0}{10.0}{10.0} & \supb{66.7}{13.3}{14.4}
             & \supv{10.7}{0.7}{0.6} & \supv{10.9}{0.7}{0.6} & \supv{14.1}{0.7}{0.6} & \supv{10.0}{0.5}{0.5} & \supb{11.7}{0.6}{0.6} & \supv{12.6}{0.9}{0.8}\\
        \bottomrule
    \end{tabular}
\end{table*}

\begin{table*}[t]
    \centering
    \caption{Scene-wise success rates (\%, left) and mean push steps over
    successful trials (right) for different methods.}
    \label{tab:app-method-combined}
    \vspace{-7pt}
    \footnotesize
    \setlength{\tabcolsep}{4pt}
    \renewcommand{\arraystretch}{1.08}
    \begin{tabular}{@{}l*{6}{c}@{\hspace{15pt}}*{6}{c}@{}}
        \toprule
        & \multicolumn{6}{c}{Success rate (\%) $\uparrow$}
        & \multicolumn{6}{c}{Push steps $\downarrow$}\\
        \cmidrule(r{15pt}){2-7}\cmidrule(l{0pt}){8-13}
        Method & S1 & S2 & S3 & S4 & S5 & S6
               & S1 & S2 & S3 & S4 & S5 & S6\\
        \midrule
        \multicolumn{13}{@{}l}{\textit{All Objects}}\\
        \addlinespace[4pt]
        Ours-100 & \supb{87.4}{5.2}{6.0} & \supb{87.1}{4.8}{5.5} & \supb{84.4}{6.1}{7.3} & \supb{84.2}{4.7}{5.4} & \supb{76.7}{6.5}{7.1} & \supb{69.4}{8.5}{9.2}
                 & \supb{10.0}{0.3}{0.3} & \supb{10.0}{0.3}{0.3} & \supb{13.3}{0.5}{0.4} & \supb{9.8}{0.3}{0.3} & \supb{11.5}{0.4}{0.4} & \supb{12.4}{0.5}{0.5}\\
        \addlinespace[4pt]
        Ours-10 & \supv{79.8}{7.1}{7.6} & \supv{79.9}{7.3}{7.9} & \supv{74.6}{8.6}{9.4} & \supv{77.6}{6.9}{7.4} & \supv{60.1}{11.2}{11.3} & \supv{53.3}{11.4}{11.7}
                & \supv{10.7}{0.5}{0.4} & \supv{10.6}{0.5}{0.4} & \supv{14.0}{0.7}{0.7} & \supv{10.4}{0.5}{0.5} & \supv{12.3}{0.7}{0.6} & \supv{12.7}{0.6}{0.6}\\
        \addlinespace[4pt]
        GRP & \supv{65.2}{10.4}{11.4} & \supv{61.8}{10.7}{10.7} & \supv{56.1}{10.6}{11.2} & \supv{58.5}{10.8}{11.3} & \supv{49.8}{13.2}{13.4} & \supv{44.5}{13.3}{13.4}
            & \supv{11.0}{0.6}{0.6} & \supv{11.0}{0.6}{0.6} & \supv{14.4}{0.7}{0.7} & \supv{10.7}{0.5}{0.5} & \supv{12.1}{0.6}{0.6} & \supv{12.9}{0.8}{0.7}\\
        \addlinespace[4pt]
        P-Net & \supv{32.5}{13.9}{13.0} & \supv{39.6}{13.0}{12.5} & \supv{17.7}{9.4}{8.2} & \supv{21.5}{12.4}{10.8} & \supv{7.1}{7.2}{5.5} & \supv{11.0}{9.7}{7.8}
              & \supv{16.2}{1.4}{1.4} & \supv{14.4}{1.3}{1.2} & \supv{19.2}{1.7}{1.8} & \supv{17.4}{2.1}{2.0} & \supv{15.0}{2.5}{2.2} & \supv{17.5}{2.1}{2.0}\\
        \addlinespace[4pt]
        ManiFM & \supv{68.2}{14.5}{15.7} & \supv{53.3}{16.5}{17.0} & \supv{44.3}{16.0}{16.2} & \supv{64.0}{14.5}{15.9} & \supv{54.5}{13.5}{14.3} & \supv{39.5}{16.0}{16.1}
               & \supv{10.8}{1.1}{0.9} & \supv{10.3}{0.6}{0.5} & \supv{13.6}{0.8}{0.7} & \supv{10.4}{0.9}{0.8} & \supv{12.6}{0.9}{0.8} & \supv{13.8}{1.4}{1.1}\\
        \midrule
        \multicolumn{13}{@{}l}{\textit{Seen Objects}}\\
        \addlinespace[4pt]
        Ours-100 & \supb{89.4}{6.2}{7.0} & \supb{89.4}{5.0}{6.2} & \supb{86.0}{6.8}{7.6} & \supb{88.8}{4.6}{5.6} & \supb{86.0}{4.4}{5.0} & \supb{72.6}{10.0}{10.4}
                 & \supb{10.1}{0.3}{0.3} & \supv{10.1}{0.4}{0.3} & \supv{12.8}{0.4}{0.4} & \supb{9.6}{0.3}{0.3} & \supb{11.3}{0.4}{0.4} & \supb{12.1}{0.3}{0.3}\\
        \addlinespace[4pt]
        Ours-10 & \supv{85.2}{8.2}{10.0} & \supv{86.8}{7.4}{8.6} & \supv{85.8}{6.2}{6.4} & \supv{85.6}{6.6}{7.0} & \supv{75.8}{10.0}{9.6} & \supv{65.8}{9.6}{8.6}
                & \supv{10.3}{0.5}{0.4} & \supv{10.4}{0.6}{0.5} & \supv{13.2}{0.8}{0.7} & \supv{9.9}{0.4}{0.4} & \supv{11.6}{0.8}{0.6} & \supv{12.3}{0.7}{0.6}\\
        \addlinespace[4pt]
        GRP & \supv{65.0}{17.6}{20.0} & \supv{62.8}{18.4}{18.6} & \supv{56.2}{17.2}{17.8} & \supv{63.2}{16.4}{18.4} & \supv{56.4}{18.8}{19.8} & \supv{45.8}{21.0}{20.0}
            & \supv{10.5}{1.0}{1.0} & \supv{10.8}{1.2}{1.2} & \supv{13.6}{1.0}{1.1} & \supv{10.2}{0.6}{0.6} & \supv{11.8}{0.8}{0.8} & \supv{12.2}{0.7}{0.8}\\
        \addlinespace[4pt]
        P-Net & \supv{39.8}{22.0}{20.4} & \supv{48.0}{22.6}{22.0} & \supv{26.4}{16.0}{14.8} & \supv{30.6}{20.2}{18.8} & \supv{8.0}{9.8}{7.0} & \supv{17.2}{17.0}{14.6}
              & \supv{16.4}{2.0}{2.2} & \supv{14.6}{2.0}{1.7} & \supv{18.9}{2.5}{2.4} & \supv{16.9}{3.0}{2.8} & \supv{13.5}{3.4}{2.4} & \supv{16.1}{1.9}{1.6}\\
        \addlinespace[4pt]
        ManiFM & \supv{72.8}{18.8}{22.2} & \supv{53.8}{23.4}{23.8} & \supv{41.8}{24.4}{22.6} & \supv{70.2}{18.2}{20.4} & \supv{57.0}{18.6}{20.2} & \supv{38.6}{24.8}{23.2}
               & \supv{10.2}{0.9}{0.8} & \supb{9.8}{0.3}{0.3} & \supb{12.6}{0.7}{0.6} & \supv{9.8}{1.1}{0.8} & \supv{11.7}{1.1}{0.8} & \supv{12.7}{0.6}{0.7}\\
        \midrule
        \multicolumn{13}{@{}l}{\textit{Unseen Objects}}\\
        \addlinespace[4pt]
        Ours-100 & \supb{85.7}{7.8}{9.2} & \supb{85.2}{7.8}{8.7} & \supb{83.0}{9.7}{11.7} & \supb{80.3}{7.2}{8.0} & \supb{69.0}{10.0}{10.0} & \supb{66.7}{13.3}{14.4}
                 & \supb{9.9}{0.5}{0.4} & \supb{10.0}{0.5}{0.4} & \supb{13.7}{0.7}{0.7} & \supb{9.9}{0.4}{0.4} & \supb{11.7}{0.6}{0.6} & \supb{12.6}{0.9}{0.8}\\
        \addlinespace[4pt]
        Ours-10 & \supv{75.3}{10.4}{10.6} & \supv{74.2}{11.0}{11.9} & \supv{65.3}{13.5}{14.0} & \supv{71.0}{10.8}{10.8} & \supv{47.0}{16.3}{15.3} & \supv{42.8}{18.9}{18.0}
                & \supv{11.1}{0.7}{0.7} & \supv{10.8}{0.7}{0.7} & \supv{14.6}{1.0}{1.0} & \supv{10.8}{0.8}{0.8} & \supv{12.8}{1.0}{0.9} & \supv{13.1}{0.9}{0.8}\\
        \addlinespace[4pt]
        GRP & \supv{65.3}{12.4}{13.0} & \supv{61.0}{12.5}{13.0} & \supv{56.0}{13.8}{14.2} & \supv{54.5}{14.2}{14.3} & \supv{44.3}{18.5}{18.1} & \supv{43.3}{18.5}{18.6}
            & \supv{11.4}{0.6}{0.6} & \supv{11.2}{0.5}{0.6} & \supv{15.1}{0.7}{0.8} & \supv{11.1}{0.7}{0.7} & \supv{12.3}{0.7}{0.8} & \supv{13.6}{1.1}{1.1}\\
        \addlinespace[4pt]
        P-Net & \supv{26.5}{18.0}{15.3} & \supv{32.7}{14.0}{13.0} & \supv{10.5}{9.0}{6.8} & \supv{14.0}{14.8}{10.0} & \supv{6.3}{10.2}{5.8} & \supv{5.8}{9.7}{5.3}
              & \supv{16.0}{1.8}{1.9} & \supv{14.3}{1.5}{1.5} & \supv{19.4}{2.4}{2.5} & \supv{18.0}{2.9}{2.7} & \supv{16.5}{3.5}{3.0} & \supv{18.9}{3.1}{3.5}\\
        \addlinespace[4pt]
        ManiFM & \supv{64.3}{21.5}{22.6} & \supv{52.8}{23.4}{23.8} & \supv{46.3}{22.0}{21.8} & \supv{58.8}{22.4}{22.8} & \supv{52.5}{19.7}{20.3} & \supv{40.2}{22.0}{21.2}
               & \supv{11.4}{1.9}{1.4} & \supv{10.8}{0.9}{0.8} & \supv{14.4}{1.1}{1.1} & \supv{11.0}{1.3}{1.1} & \supv{13.3}{1.3}{1.2} & \supv{14.7}{2.3}{1.8}\\
        \bottomrule
    \end{tabular}
\end{table*}

\subsection{Details of Baseline Setting}
\label{app:baselines}

\textbf{GRP:} the Greedy Reactive Policy is a heuristic method designed in this work.
Its 2D version follows~\cite{Li2018PushNet}.
There, we extend it to 3D and enhance it with our stability evaluation model.
It sets the magnitude of the contact vector equal to the expected translational norm.
The contact direction is obtained by rotating the opposite direction of
the translation component of the expected transform by its rotational angle.
Candidate contact points are then selected from points on the object
whose projections onto the $xy$-plane form an angle smaller than $20^\circ$ with
the opposite translation direction and pass through the centroid. 
Finally, the stability model evaluates and selects the contact point.

\textbf{P-Net:} P-Net is an enhanced version of the original PushNet.
The original PushNet performs planar pushing using a single-view camera,
which limites viewpoint coverage and makes data collection cumbersome.
We incorporate an object-centric design by reconstructing the point cloud from three camera views
and projecting it into a 2D mask as input.
In addition, both the image and the action are rotated in an object-centric manner 
to leverage our single-direction contact dataset.
During action execution, a bottom-pushing strategy is adopted.

\begin{table*}[t]
    \centering
    \caption{Scene-wise success rates (\%, left) and mean push steps over
    successful trials (right) for the ablation study.}
    \label{tab:app-ablation-combined}
    \vspace{-7pt}
    \footnotesize
    \setlength{\tabcolsep}{4pt}
    \renewcommand{\arraystretch}{1.08}
    \begin{tabular}{@{}l*{6}{c}@{\hspace{15pt}}*{6}{c}@{}}
        \toprule
        & \multicolumn{6}{c}{Success rate (\%) $\uparrow$}
        & \multicolumn{6}{c}{Push steps $\downarrow$}\\
        \cmidrule(r{15pt}){2-7}\cmidrule(l{0pt}){8-13}
        Method & S1 & S2 & S3 & S4 & S5 & S6
               & S1 & S2 & S3 & S4 & S5 & S6\\
        \midrule
        \multicolumn{13}{@{}l}{\textit{All Objects}}\\
        \addlinespace[4pt]
        Ours & \supb{87.4}{5.2}{6.0} & \supb{87.1}{4.8}{5.5} & \supb{84.4}{6.1}{7.3} & \supb{84.2}{4.7}{5.4} & \supb{76.7}{6.5}{7.1} & \supb{69.4}{8.5}{9.2}
             & \supv{10.0}{0.3}{0.3} & \supv{10.0}{0.3}{0.3} & \supv{13.3}{0.5}{0.4} & \supv{9.8}{0.3}{0.3} & \supv{11.5}{0.4}{0.4} & \supv{12.4}{0.5}{0.5}\\
        \addlinespace[4pt]
        \(-\)Stab & \supv{83.4}{5.7}{6.0} & \supv{81.3}{6.9}{7.8} & \supv{79.1}{7.4}{8.3} & \supv{81.5}{4.8}{5.3} & \supv{75.2}{6.3}{6.9} & \supv{68.0}{8.0}{8.7}
                   & \supv{10.1}{0.4}{0.3} & \supv{10.0}{0.3}{0.3} & \supv{13.2}{0.4}{0.4} & \supv{9.8}{0.3}{0.3} & \supv{11.6}{0.4}{0.4} & \supv{12.5}{0.5}{0.4}\\
        \addlinespace[4pt]
        Stab only & \supv{8.9}{10.4}{6.7} & \supv{9.3}{9.9}{6.3} & \supv{6.5}{9.2}{5.4} & \supv{7.2}{9.9}{5.8} & \supv{4.3}{8.2}{4.3} & \supv{3.2}{6.1}{3.2}
                  & \supv{13.2}{1.9}{1.7} & \supv{14.3}{1.4}{1.5} & \supv{15.0}{1.7}{1.7} & \supv{12.6}{1.7}{1.7} & \supv{14.0}{5.0}{3.9} & \supv{11.5}{0.9}{0.9}\\
        \addlinespace[4pt]
        \(-\)Cost & \supv{81.0}{5.6}{5.9} & \supv{80.5}{6.7}{7.4} & \supv{74.2}{8.0}{8.7} & \supv{81.7}{5.8}{6.5} & \supv{76.4}{6.9}{7.6} & \supv{22.5}{9.7}{8.4}
                   & \supb{9.8}{0.3}{0.3} & \supv{9.7}{0.3}{0.3} & \supv{12.8}{0.4}{0.4} & \supv{9.6}{0.2}{0.2} & \supb{11.1}{0.3}{0.2} & \supv{10.6}{0.5}{0.5}\\
        \addlinespace[4pt]
        \(-\)Fb & \supv{76.3}{6.5}{7.1} & \supv{74.5}{6.6}{7.1} & \supv{61.8}{8.3}{9.1} & \supv{72.6}{7.3}{8.0} & \supv{51.7}{6.8}{6.6} & \supv{15.7}{6.5}{5.2}
                 & \supv{9.8}{0.2}{0.2} & \supb{9.7}{0.2}{0.2} & \supb{12.7}{0.4}{0.3} & \supb{9.5}{0.2}{0.2} & \supv{11.1}{0.4}{0.3} & \supb{10.3}{0.6}{0.7}\\
        \midrule
        \multicolumn{13}{@{}l}{\textit{Seen Objects}}\\
        \addlinespace[4pt]
        Ours & \supb{89.4}{6.2}{7.0} & \supb{89.4}{5.0}{6.2} & \supb{86.0}{6.8}{7.6} & \supb{88.8}{4.6}{5.6} & \supb{86.0}{4.4}{5.0} & \supb{72.6}{10.0}{10.4}
             & \supv{10.1}{0.3}{0.3} & \supv{10.1}{0.4}{0.3} & \supv{12.8}{0.4}{0.4} & \supv{9.6}{0.3}{0.3} & \supv{11.3}{0.4}{0.4} & \supv{12.1}{0.3}{0.3}\\
        \addlinespace[4pt]
        \(-\)Stab & \supv{88.2}{6.2}{6.6} & \supv{85.4}{7.6}{8.2} & \supv{85.8}{6.8}{7.4} & \supv{85.4}{5.4}{6.0} & \supv{81.8}{6.6}{6.2} & \supv{72.6}{10.4}{11.4}
                   & \supv{10.1}{0.4}{0.3} & \supv{10.0}{0.4}{0.3} & \supv{12.8}{0.3}{0.3} & \supv{9.6}{0.2}{0.2} & \supv{11.3}{0.4}{0.4} & \supv{12.3}{0.5}{0.5}\\
        \addlinespace[4pt]
        Stab only & \supv{15.4}{20.6}{12.6} & \supv{15.4}{19.8}{12.2} & \supv{12.6}{19.0}{10.8} & \supv{13.0}{20.2}{11.6} & \supv{9.4}{18.2}{9.4} & \supv{7.0}{13.4}{7.0}
                  & \supv{12.1}{2.2}{1.6} & \supv{14.1}{2.4}{2.4} & \supv{14.7}{2.1}{1.9} & \supv{12.3}{2.2}{2.1} & \supv{14.0}{5.0}{3.9} & \supv{11.5}{0.9}{0.9}\\
        \addlinespace[4pt]
        \(-\)Cost & \supv{84.2}{7.8}{7.8} & \supv{85.8}{6.4}{7.6} & \supv{81.2}{8.2}{8.6} & \supv{86.6}{6.2}{7.4} & \supv{82.6}{7.0}{7.2} & \supv{19.6}{12.2}{9.6}
                   & \supb{9.9}{0.3}{0.3} & \supb{9.8}{0.5}{0.4} & \supv{12.4}{0.3}{0.3} & \supv{9.5}{0.3}{0.3} & \supv{10.9}{0.3}{0.3} & \supv{10.1}{0.7}{0.7}\\
        \addlinespace[4pt]
        \(-\)Fb & \supv{81.6}{8.0}{7.8} & \supv{80.0}{7.4}{7.8} & \supv{71.6}{6.8}{7.2} & \supv{81.2}{6.4}{5.4} & \supv{55.0}{8.0}{7.8} & \supv{14.8}{5.4}{4.6}
                 & \supv{9.9}{0.3}{0.3} & \supv{9.8}{0.3}{0.3} & \supb{12.4}{0.3}{0.3} & \supb{9.5}{0.3}{0.2} & \supb{10.9}{0.3}{0.3} & \supb{9.8}{0.7}{0.7}\\
        \midrule
        \multicolumn{13}{@{}l}{\textit{Unseen Objects}}\\
        \addlinespace[4pt]
        Ours & \supb{85.7}{7.8}{9.2} & \supb{85.2}{7.8}{8.7} & \supb{83.0}{9.7}{11.7} & \supb{80.3}{7.2}{8.0} & \supv{69.0}{10.0}{10.0} & \supb{66.7}{13.3}{14.4}
             & \supv{9.9}{0.5}{0.4} & \supv{10.0}{0.5}{0.4} & \supv{13.7}{0.7}{0.7} & \supv{9.9}{0.4}{0.4} & \supv{11.7}{0.6}{0.6} & \supv{12.6}{0.9}{0.8}\\
        \addlinespace[4pt]
        \(-\)Stab & \supv{79.3}{8.7}{8.6} & \supv{77.8}{10.9}{11.3} & \supv{73.5}{11.7}{12.3} & \supv{78.2}{7.5}{7.7} & \supv{69.7}{10.1}{10.4} & \supv{64.2}{11.8}{12.5}
                   & \supv{10.2}{0.6}{0.5} & \supv{10.0}{0.5}{0.4} & \supv{13.5}{0.6}{0.6} & \supv{10.0}{0.4}{0.4} & \supv{11.8}{0.7}{0.6} & \supv{12.6}{0.7}{0.7}\\
        \addlinespace[4pt]
        Stab only & \supv{3.5}{5.2}{3.2} & \supv{4.2}{4.5}{3.5} & \supv{1.3}{2.4}{1.3} & \supv{2.3}{2.9}{2.1} & \supv{0.0}{0.0}{0.0} & \supv{0.0}{0.0}{0.0}
                  & \supv{15.6}{2.5}{2.1} & \supv{14.5}{1.0}{1.5} & \supv{16.1}{1.9}{1.9} & \supv{13.2}{2.3}{2.2} & -- & --\\
        \addlinespace[4pt]
        \(-\)Cost & \supv{78.3}{7.5}{8.1} & \supv{76.0}{10.3}{11.5} & \supv{68.3}{12.5}{13.3} & \supv{77.7}{9.0}{9.0} & \supb{71.2}{10.5}{11.5} & \supv{25.0}{15.0}{12.3}
                   & \supb{9.7}{0.4}{0.4} & \supv{9.7}{0.4}{0.4} & \supv{13.0}{0.6}{0.6} & \supv{9.6}{0.3}{0.3} & \supb{11.2}{0.4}{0.4} & \supv{11.0}{0.5}{0.6}\\
        \addlinespace[4pt]
        \(-\)Fb & \supv{71.8}{9.4}{10.5} & \supv{69.8}{9.4}{10.5} & \supv{53.7}{12.5}{13.7} & \supv{65.5}{11.3}{12.0} & \supv{49.0}{10.5}{9.7} & \supv{16.5}{10.8}{8.7}
                 & \supv{9.8}{0.4}{0.3} & \supb{9.5}{0.3}{0.2} & \supb{13.0}{0.6}{0.5} & \supb{9.5}{0.4}{0.3} & \supv{11.2}{0.6}{0.5} & \supb{10.8}{0.9}{1.0}\\
        \bottomrule
    \end{tabular}
\end{table*}

\textbf{Aug ManiFM:} 
ManiFM is also an enhanced version of the original ManiFM.
We augment the original model by introducing an MLP with 
layer normalization to extract point-cloud motion features,
and replace the MSE loss with a BCE loss to address the sparsity of contact points.
Meanwhile, since the original ManiFM takes the robot body point cloud as input, 
we decouple it from the policy and keep it fixed, 
such that it no longer affects the affordance of the object itself.
In addition, we set the expected translation norm of the object as the contact distance.
Finally, object-centric rotation augmentation is applied to 
the data to utilize our single-direction contact dataset.

\textbf{CEM-MPC:} 
CEM-MPC is a model-based baseline that uses MuJoCo~\cite{todorov2012mujoco} 
as a black-box dynamics model. 
Given the reconstructed object geometry, current pose, 
and target sub-goal, it samples 512 surface contacts and optimizes push contact, direction, 
and distance using CEM. The push direction varies within $\pm 90^\circ$ around the inward surface normal, 
and the pushing distance is limited to $0.1\,\mathrm{m}$.

CEM runs for 12 iterations with 1,024 candidate actions per iteration. 
After uniform initialization, the best (10\%) of samples are used to refit the sampling distribution, 
while (20\%) uniform samples are retained for exploration. 
Candidate pushes are evaluated through MuJoCo rollouts using a cost 
that penalizes goal error, yaw error, tipping, excessive vertical motion, workspace violations, 
and push distance. Invalid rollouts receive a large penalty. 
The lowest-cost action is converted into end-effector start and end poses and executed by the motion planner.

\subsection{Supplementary Results with Different Horizons}
\label{app:horizons}

Table~\ref{tab:app-horizon-combined} jointly reports the scene-wise success
rates (left) and push-step counts (right) for different horizon lookahead. 
For each metric, we report results over all 22
objects and separately over the 10 seen and 12 unseen objects. 
An entry \(m^{+\Delta_+}_{-\Delta_-}\) denotes the mean and its asymmetric 95\%
bootstrap confidence interval over objects; boldface marks the highest success
rate or the lowest push-step mean in each scene.
A \emph{push step} is one push primitive physically executed.
For every object, we first average the number of executed pushes over successful trials, 
and then average these object-level means. 

\noindent\textbf{Analysis.}
One-step lookahead gives the highest success rates in Scenes~1--4, whereas
full-horizon lookahead is most effective in the two adversarial scenes. The
gain is especially large in Scene~6: the all-object success rate rises from
30.1\% at \(h=1\) to 69.4\% with the full horizon. The same trend holds for
seen objects (27.8\% to 72.6\%) and unseen objects (32.0\% to 66.7\%).
Longer lookahead can reject a locally attractive route before it
enters a region that is difficult for the robot to execute. 
However, in simpler scenes, such additional anticipation is detrimental.

The pushing step counts vary much less than the success rates.
One-step lookahead is shortest in Scenes~1--4, while the full horizon
is slightly shorter in Scene~5 and simultaneously succeeds more often. 
In Scene~6, full-horizon planning substantially improves the success rate 
with only a minor increase in the number of steps. 
This trend is consistent across both seen and unseen objects. 
Overall, longer horizons tend to perform better in challenging scenarios involving difficult choices, 
whereas in simpler scenes they may introduce redundant information and degrade performance.

\begin{table*}[t]
    \centering
    \caption{Scene-wise success rates (\%, left) and mean push steps over
    successful trials (right) with estimated and ground-truth object poses.}
    \label{tab:app-pose-estimation}
    \vspace{-7pt}
    \footnotesize
    \setlength{\tabcolsep}{4pt}
    \renewcommand{\arraystretch}{1.08}
    \begin{tabular}{@{}l*{6}{c}@{\hspace{15pt}}*{6}{c}@{}}
        \toprule
        & \multicolumn{6}{c}{Success rate (\%) $\uparrow$}
        & \multicolumn{6}{c}{Push steps $\downarrow$}\\
        \cmidrule(r{15pt}){2-7}\cmidrule(l{0pt}){8-13}
        Pose & S1 & S2 & S3 & S4 & S5 & S6
             & S1 & S2 & S3 & S4 & S5 & S6\\
        \midrule
        \multicolumn{13}{@{}l}{\textit{All Objects}}\\
        \addlinespace[4pt]
        Est. & \supv{87.4}{5.2}{6.0} & \supv{87.1}{4.8}{5.5} & \supv{84.4}{6.1}{7.3} & \supv{84.2}{4.7}{5.4} & \supv{76.7}{6.5}{7.1} & \supv{69.4}{8.5}{9.2}
             & \supb{10.0}{0.3}{0.3} & \supv{10.0}{0.3}{0.3} & \supv{13.3}{0.5}{0.4} & \supv{9.8}{0.3}{0.3} & \supv{11.5}{0.4}{0.4} & \supv{12.4}{0.5}{0.5}\\
        \addlinespace[4pt]
        GT & \supb{89.5}{3.6}{4.0} & \supb{88.1}{4.3}{4.8} & \supb{84.4}{5.6}{6.6} & \supb{85.6}{4.1}{4.5} & \supb{78.4}{7.3}{8.8} & \supb{71.2}{7.8}{8.7}
           & \supv{10.1}{0.3}{0.3} & \supb{10.0}{0.2}{0.2} & \supb{13.1}{0.3}{0.3} & \supb{9.7}{0.3}{0.2} & \supb{11.4}{0.3}{0.3} & \supb{12.2}{0.3}{0.3}\\
        \midrule
        \multicolumn{13}{@{}l}{\textit{Seen Objects}}\\
        \addlinespace[4pt]
        Est. & \supb{89.4}{6.2}{7.0} & \supv{89.4}{5.0}{6.2} & \supv{86.0}{6.8}{7.6} & \supv{88.8}{4.6}{5.6} & \supb{86.0}{4.4}{5.0} & \supv{72.6}{10.0}{10.4}
             & \supb{10.1}{0.3}{0.3} & \supv{10.1}{0.4}{0.3} & \supv{12.8}{0.4}{0.4} & \supv{9.6}{0.3}{0.3} & \supv{11.3}{0.4}{0.4} & \supb{12.1}{0.3}{0.3}\\
        \addlinespace[4pt]
        GT & \supv{89.0}{6.2}{6.2} & \supb{90.4}{5.0}{5.0} & \supb{87.8}{5.4}{5.6} & \supb{90.0}{4.2}{4.4} & \supv{85.4}{6.4}{7.4} & \supb{76.0}{9.0}{9.2}
           & \supv{10.2}{0.4}{0.3} & \supb{10.0}{0.3}{0.3} & \supb{12.7}{0.2}{0.3} & \supb{9.6}{0.2}{0.2} & \supb{11.3}{0.3}{0.3} & \supv{12.1}{0.4}{0.4}\\
        \midrule
        \multicolumn{13}{@{}l}{\textit{Unseen Objects}}\\
        \addlinespace[4pt]
        Est. & \supv{85.7}{7.8}{9.2} & \supv{85.2}{7.8}{8.7} & \supb{83.0}{9.7}{11.7} & \supv{80.3}{7.2}{8.0} & \supv{69.0}{10.0}{10.0} & \supv{66.7}{13.3}{14.4}
             & \supb{9.9}{0.5}{0.4} & \supv{10.0}{0.5}{0.4} & \supv{13.7}{0.7}{0.7} & \supv{9.9}{0.4}{0.4} & \supv{11.7}{0.6}{0.6} & \supv{12.6}{0.9}{0.8}\\
        \addlinespace[4pt]
        GT & \supb{89.8}{4.0}{5.3} & \supb{86.2}{6.3}{7.9} & \supv{81.5}{8.8}{10.7} & \supb{82.0}{6.0}{6.5} & \supb{72.5}{11.5}{13.7} & \supb{67.2}{11.8}{13.4}
           & \supv{10.0}{0.5}{0.4} & \supb{10.0}{0.4}{0.3} & \supb{13.3}{0.5}{0.5} & \supb{9.8}{0.4}{0.4} & \supb{11.5}{0.5}{0.4} & \supb{12.2}{0.5}{0.5}\\
        \bottomrule
    \end{tabular}
\end{table*}

\subsection{Supplementary Results of Different Methods}
\label{app:methods}

Table~\ref{tab:app-method-combined} shows the comparison about our method with the baselines across six scenes 
and two object categories.

\noindent\textbf{Analysis.}
Ours-100 attains the highest success rate in every scene and the lowest
all-object push-step mean in every scene. 
GRP relies on heuristic design and struggles to handle diverse 3D objects. 
PushNet is primarily designed for 2D convex objects and therefore also performs poorly in this setting. 
ManiFM achieves competitive push-step counts on seen objects in Scenes~2 and~3, 
but remains close to our method. 
In other settings, especially in success rate, it shows a large gap, indicating lower robustness.
The results of Ours-10 further demonstrate strong generalization of our method across object shapes. 
Although reducing training-set diversity degrades both success rate and step efficiency, 
Ours-10 still performs competitively against other methods and shows only a moderate drop relative to Ours-100.

\subsection{Supplementary Ablation Study}
\label{app:Ablation}

Table~\ref{tab:app-ablation-combined} jointly reports scene-wise success rates
(left) and successful-trial push-step means (right) for the four ablations: 
\textit{\(-\)Stab} removes stability scoring;
\textit{Stab only} selects contacts using only that score; \textit{\(-\)Cost}
disables local path-cost updates after motion-planning failures; and
\textit{\(-\)Fb} removes the future feasibility check and its feedback. 
Each metric is reported over all, seen, and unseen objects, with the same
confidence-interval convention as above.

\noindent\textbf{Analysis.}
In terms of success rate, the complete method achieves the best overall performance across all scenes. 
The \textit{\(-\)Cost} variant performs relatively well in Scene~5, 
where feasibility failures trigger stochastic replanning without conveying failure-specific information. 
Because the planner does not strongly favor paths near the infeasible in this scene, 
repeated replanning can still eventually find a feasible route.
For the number of steps, \textit{\(-\)Fb} and \textit{\(-\)Cost} perform slightly better because, 
without additional cost penalties, the planner tends to choose shorter paths to
the goal. However, these minor gains are outweighed by the
substantial decrease in success rate.

\subsection{Analysis of Pose Estimation Impact}
\label{app:pose-estimation}

In simulation, we reconstruct the object point cloud from three cameras and
estimate its pose using the PCA--ICP pipeline described in
Sec.~\ref{sec:new_experiments}. To isolate the influence of pose estimation,
we additionally evaluate Ours-100 using the ground-truth (GT) object pose while
keeping the scenes, trials, horizon assignment, and all other components
unchanged. Table~\ref{tab:app-pose-estimation} jointly compares the success
rates and push-step counts obtained with estimated and GT poses. Here,
\emph{Est.} denotes PCA--ICP pose estimation.

Using ground-truth poses improves both overall task success and push efficiency, but only marginally. 
This is because PCA-ICP with consecutive-frame constraints provides accurate pose estimates in most cases. 
For fully or nearly symmetric objects, such as cylinders, 
orientation ambiguity does not affect the Chamfer-distance-based success criterion 
or the computation of single-step transformations.

For the real-world experiments in Section~\ref{sec:real-world}, 
a major source of failure is perception error. 
Unlike simulation, where the point cloud is reconstructed first and the object pose is then estimated, 
the real-world setup follows the opposite pipeline. With only a single RGB-D camera, 
FoundationPose first estimates the object pose, 
which is then used to transform the reference point cloud to the current pose before prediction. 
Consequently, pose estimation errors can cause substantial misalignment between the transformed point cloud 
and the true object surface, leading to inaccurate push predictions and large end-effector execution errors.

\end{document}